\documentclass{article}

\usepackage[final]{neurips_2020}

\usepackage[utf8]{inputenc} %
\usepackage[T1]{fontenc}    %
\usepackage{hyperref}       %
\usepackage{url}            %
\usepackage{booktabs}       %
\usepackage{amsfonts}       %
\usepackage{nicefrac}       %
\usepackage{microtype}      %
\usepackage{algorithm}%
\usepackage{algpseudocode}%
\usepackage{microtype}
\usepackage{graphicx}
\usepackage{subcaption}
\usepackage{booktabs} %
\usepackage{multicol}
\usepackage{graphicx}
\usepackage{hyperref}
\usepackage{todonotes}
\usepackage{url}
\usepackage{bbm}
\usepackage{float} 
\usepackage{wrapfig}
\usepackage{enumitem}

\newcommand\numberthis{\addtocounter{equation}{1}\tag{\theequation}}

\usepackage{amsmath,amsthm,amsfonts,bm}

\newtheorem{theorem}{Theorem}[section]

\def\eqref#1{equation~\ref{#1}}

\def\1{\bm{1}}

\DeclareMathAlphabet{\mathsfit}{\encodingdefault}{\sfdefault}{m}{sl}
\SetMathAlphabet{\mathsfit}{bold}{\encodingdefault}{\sfdefault}{bx}{n}

\def\gA{{\mathcal{A}}}

\def\gD{{\mathcal{D}}}

\def\gM{{\mathcal{M}}}

\def\gS{{\mathcal{S}}}
\def\gT{{\mathcal{T}}}

\newcommand{\E}{\mathbb{E}}

\DeclareMathOperator*{\argmax}{arg\,max}

\newcommand{\goalfn}{\mathcal{G}}
\newcommand{\oldpi}{\pi_{old}}

\title{Learning to Reach Goals via \\Iterated Supervised Learning}

\author{%
  Dibya Ghosh\thanks{First two authors contributed equally. Correspondence at \texttt{dibya.ghosh@berkeley.edu}} \\
  UC Berkeley\\
   \And
   Abhishek Gupta$^*$ \\
   UC Berkeley \\
   \And
   Ashwin Reddy \\
   UC Berkeley \\
      \And
   Justin Fu \\
   UC Berkeley \\
      \AND
   Coline Devin \\
   UC Berkeley \\
      \And
   Benjamin Eysenbach \\
   Carnegie Mellon University \\
      \And
   Sergey Levine \\
   UC Berkeley \\
}

\begin{document}

\maketitle

\begin{abstract}
Current reinforcement learning (RL) algorithms can be brittle and difficult to use, especially when learning goal-reaching behaviors from sparse rewards. Although supervised imitation learning provides a simple and stable alternative, it requires access to demonstrations from a human supervisor. In this paper, we study RL algorithms that use imitation learning to acquire goal reaching policies from scratch, without the need for expert demonstrations or a value function. In lieu of demonstrations, we leverage the property that any trajectory is a successful demonstration for reaching the final state in that same trajectory. We propose a simple algorithm in which an agent continually relabels and imitates the trajectories it generates to progressively learn goal-reaching behaviors from scratch. Each iteration, the agent collects new trajectories using the latest policy, and maximizes the likelihood of the actions along these trajectories under the goal that was actually reached, so as to improve the policy. We formally show that this iterated supervised learning procedure optimizes a bound on the RL objective, derive performance bounds of the learned policy, and empirically demonstrate improved goal-reaching performance and robustness over current RL algorithms in several benchmark tasks. 
\end{abstract}

\section{Introduction}
\label{sec:intro}

Reinforcement learning (RL) provides an elegant framework for agents to learn general-purpose behaviors supervised by only a reward signal. When combined with neural networks, RL has enabled many notable successes, but our most successful deep RL algorithms are far from a turnkey solution. Despite striving for data efficiency, RL algorithms, especially those using temporal difference learning, are highly sensitive to hyperparameters \citep{henderson2018deep} and face challenges of stability and optimization \citep{tsitsiklis1997analysis, Hasselt2018DeepRL, kumar2019stabilizing}, making such algorithms difficult to use in practice. 

If agents are supervised not with a reward signal, but rather demonstrations from an expert, the resulting class of algorithms is significantly more stable and easy to use. Imitation learning via behavioral cloning provides a simple paradigm for training control policies: maximizing the likelihood of optimal actions via supervised learning. Imitation learning algorithms using deep learning are mature and robust; these algorithms have demonstrated success in acquiring behaviors reliably from high-dimensional sensory data such as images~\citep{bojarski2016imitation, lynch2019learning}. Although imitation learning via supervised learning is not a replacement for RL -- the paradigm is limited by the difficulty of obtaining kinesthetic demonstrations from a supervisor -- the idea of learning policies via supervised learning can serve as inspiration for RL agents that learn behaviors from scratch.

In this paper, we present a simple RL algorithm for learning goal-directed policies that leverages the stability of supervised imitation learning without requiring an expert supervisor. 
We show that when learning goal-directed behaviors using RL, demonstrations of optimal behavior can be generated from sub-optimal data in a fully self-supervised manner using the principle of data relabeling: that every trajectory is a successful demonstration for the state that it \emph{actually} reaches, even if it is sub-optimal for the goal that was originally commanded to generate the trajectory. A similar observation of hindsight relabelling was originally made by \citet{kaelbling1993goals}, more recently popularized in the deep RL literature~\citep{andrychowicz2017her}, for learning with off-policy value-based methods and policy-gradient methods \citep{rauber2017hindsight}. 
When goal-relabelling, these algorithms recompute the received rewards as though a different goal had been commanded. In this work, we instead notice that goal-relabelling to the final state in the trajectory allows an algorithm to re-interpret an action collected by a sub-optimal agent as though it were collected by an expert agent, just for a different goal. %
This leads to a substantially simpler algorithm that relies only on a supervised imitation learning primitive, avoiding the challenges of value function estimation. By generating demonstrations using hindsight relabelling, we are able to apply goal-conditioned imitation learning primitives \citep{gupta2019relay, ding2019goal} on data collected by \textit{sub-optimal} agents, not just from an expert supervisor.%

We instantiate these ideas as an algorithm that we call goal-conditioned supervised learning (GCSL). At each iteration, trajectories are collected commanding the current goal-conditioned policy for some set of desired goals, and then relabeled using hindsight to be optimal for the set of goals that were actually reached. Supervised imitation learning with this generated ``expert'' data is used to train an improved goal-conditioned policy for the next iteration. Interestingly, this simple procedure provably optimizes a lower bound on a well-defined RL objective; by performing self-imitation on all of its own trajectories, an agent can iteratively improve its own policy to learn optimal goal-reaching behaviors without requiring any external demonstrations and without learning a value function. While self-imitation RL algorithms typically choose a small subset of trajectories to imitate~\citep{oh2018sil, Hao2019IndependentGA} or learn a separate value function to reweight past experience~\citep{neumann2009fitted,abdolmaleki2018maximum,  peng2019advantage}, we show that GCSL learns efficiently while training on \emph{every} previous trajectory without reweighting, thereby maximizing data~reuse. %

The main contribution of our work is GCSL, a \textit{simple} goal-reaching RL algorithm that uses supervised learning to acquire policies from scratch. We show, both formally and empirically, that \textit{any} trajectory taken by the agent can be turned into an optimal one using hindsight relabelling, and that imitation of these trajectories (provably) enables an agent to (iteratively) learn goal-reaching behaviors. That iteratively imitating all the data from a sub-optimal agent leads to optimal behavior is a non-trivial conclusion; we formally verify that the procedure optimizes a lower-bound on a goal-reaching RL objective and derive performance bounds when the supervised learning objective is sufficiently minimized. 
In practice, GCSL is simpler, more stable, and less sensitive to hyperparameters than value-based methods, while still retaining the benefits of off-policy learning. Moreover, GCSL can leverage demonstrations (if available) to accelerate learning. We demonstrate that GCSL outperforms value-based and policy gradient methods on several challenging robotic domains.%

\section{Preliminaries}
\label{sec:background}

\textbf{Goal reaching.} The goal reaching problem is characterized by the tuple $\langle \gS, \gA, \gT, \rho(s_0), T, p(g)\rangle$, where $\gS$ and $\gA$ are the state and action spaces, $\gT(s' |s,a)$ is the transition function, $\rho(s_0)$ is the initial state distribution, $T$ the horizon length, and $p(g)$ is the distribution over goal states $g \in \gS$. We aim to find a time-varying goal-conditioned policy $\pi(\cdot | s, g, h)$: $\gS \times \gS \times [T] \to \Delta(\gA)$, where $\Delta(\gA)$ is the probability simplex over the action space $\gA$ and $h$ is the remaining horizon. We will say that a goal is achieved if the agent has reached the goal at the end of the episode.
Correspondingly, the learning problem is to acquire a policy that maximizes the probability of achieving the desired goal:
\begin{equation}
J(\pi) = \mathbb{E}_{g \sim p(g)} \left[P_{\pi_g}\left(s_T = g\right) \right]. \label{eq:main-reward}
\end{equation}
Notice that unlike a shortest-path objective, this objective provides no incentive to find the shortest path to the goal, but rather incentivizes behaviours that are more stable and safe, that are \textit{guaranteed} to reach the goal over potentially risky shorter paths. We shall see in Section \ref{sec:method}
that this notion of optimality is more than a simple design choice: hindsight relabeling for optimality emerges naturally when maximizing the probability of achieving the goal, but does not when minimizing the time to reach the goal.
 
\textbf{Goal-conditioned RL.} The goal reaching problem can be equivalently defined using the nomenclature of RL as a collection of Markov decision processes (MDPs) $\{\mathcal{M}_g\}_{g \in \gS}$. Each MDP $\gM_g$ is defined as the tuple $\langle \gS, \gA, \gT_g, r_g, \rho, T\rangle$, where the state space, action space, initial state distribution, and horizon as above. For each goal, a reward function is defined as $r_g(s) = \mathbbm{1}(s=g)$. Using this notation, an optimal goal-conditioned policy maximizes the return in an MDP $\gM_g$ sampled according to the goal distribution, 
\begin{equation}
J(\pi) = \mathbb{E}_{g \sim p(g)} \left[\mathbb{E}_{\tau \sim \pi_g}\left[ r_g(s_T)\right]\right]. \label{eq:main-reward-rl}
\end{equation}
Since the transition dynamics are equivalent for different goals, off-policy value-based methods can use transitions collected for one goal to compute the value function for arbitrary other goals. Namely, \citet{kaelbling1993goals} first showed that if the transition $(s, a, s' ,r)$ was witnessed when reaching a specific goal $g$, it can be relabeled to $(s, a, s', r_{g'}(s))$ for an arbitrary goal $g' \in \gS$ if the underlying goal reward function is known. Hindsight experience replay \citep{andrychowicz2017her} considers a specific case of relabeling to when the relabeled goal is another state further down the trajectory. 

\textbf{Goal-conditioned imitation learning.} If an agent is additionally provided expert demonstrations for reaching particular goals, behavioral cloning is a simple algorithm to learn the optimal policy by maximizing the likelihood of the demonstration data under the policy. Formally, demonstrations are provided as a dataset of expert behavior $\mathcal{D}^* = \{\tau_1, \tau_2, \dots\}$ from an expert policy $\pi^*$, where each trajectory $\tau_i = \{s_0^i, a_0^i, s_1^i, a_1^i, ....s_T^1\}$ is optimal for reaching the final state in the trajectory. Given a parametric class of stochastic, time-varying policies $\Pi$, the behavioral cloning objective is to maximize the likelihood of actions seen in the data when attempting to reach this desired goal, 
\begin{equation*}
\begin{split}
\pi_{BC} &= \argmax_{\pi \in \Pi} \mathbb{E}_{\tau \sim \pi^*} \left[ \log \pi(a_{t}|s = s_{t}, g = s_{T}, h=T-t) \right] \qquad \text{for $0 \leq t \leq T$}.
\end{split}
\end{equation*}

\section{Learning Goal-Conditioned Policies with Self-Imitation}
\label{sec:method}

\begin{figure*}[t]
    \centering
    \vspace{-2em}
    \includegraphics[width=\textwidth]{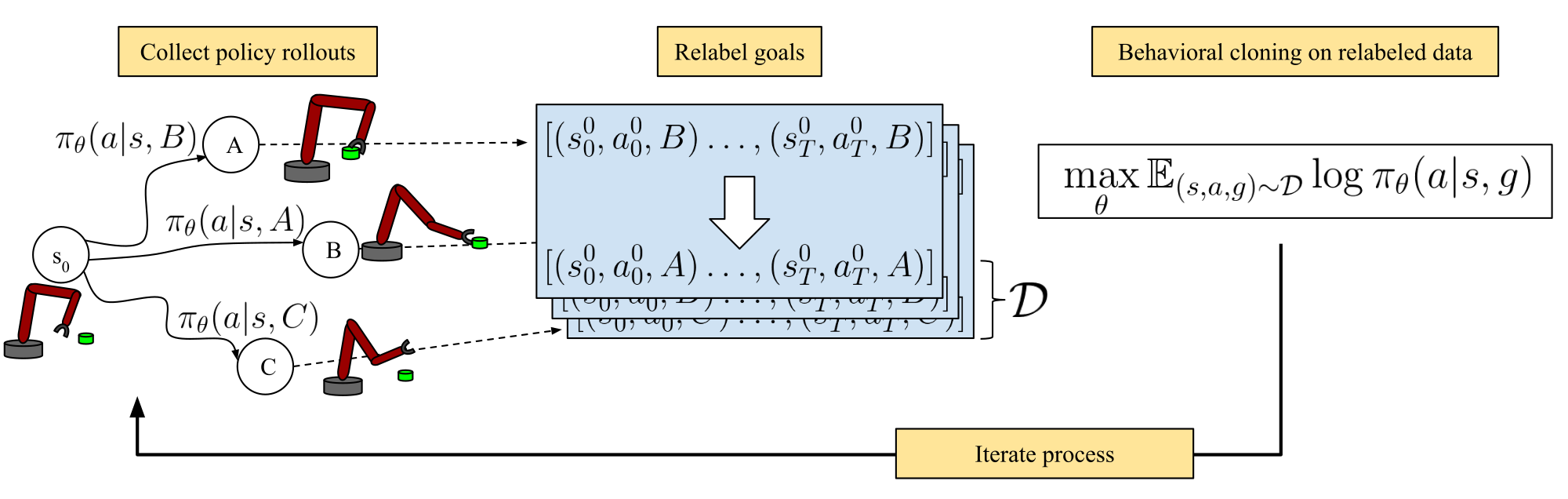}
    \vspace{-1em}
    \caption{Goal-conditioned supervised learning (GCSL): The agent learns how to reach goals by sampling trajectories, relabeling the trajectories to be optimal in hindsight and treating them as expert data, and then performing supervised learning via behavioral cloning.}
    \vspace{-1em}
    \label{fig:teaser}
\end{figure*}

In this section, we show how imitation learning via behavior cloning with data relabeling can be utilized in an iterative procedure that optimizes a lower bound on the RL objective. The resulting procedure, in which an agent continually relabels and imitates its own experience, is \emph{not} an imitation learning algorithm, but rather an RL algorithm for learning goal-reaching from scratch without any expert demonstrations. This algorithm, illustrated in Fig.~\ref{fig:teaser}, is simple and allows us to perform off-policy reinforcement learning for goal reaching without learning value functions. 

\subsection{Goal-Conditioned Supervised Learning}
We can attain the benefits of behavioral cloning without the dependence on human supervision by leveraging the following insight: under last-timestep optimality (Equation \ref{eq:main-reward}), a trajectory that fails to reach the intended goal is nonetheless optimal for reaching the goal it actually reached. As a result, a  trajectory from a sub-optimal agent can be re-interpreted by goal-conditioned behavior cloning as an optimal trajectory for reaching a potentially different goal. This insight will allow us to convert sub-optimal trajectories into optimal goal reaching trajectories for different goals, without the need for any human supervision.

More precisely, consider a trajectory $\tau = \{s_1, a_1, s_2, a_2, \dots, s_T, a_T\}$ obtained by commanding the policy  $\pi_{\theta}(a \mid s, g, h)$ to reach some goal $g$.
For any time step $t$ and horizon $h$, the action $a_t$ in state $s_t$ is {likely to be a good action} for reaching $s_{t+h}$ in $h$ time steps (even if it is not a good action for reaching the originally commanded goal $g$), and thus can be treated as expert supervision for $\pi_{\theta}(\cdot \mid s_t, s_{t+h}, h)$.  To obtain a concrete algorithm, we can relabel all time steps and horizons in a trajectory to create an expert dataset according to
\begin{equation}
\mathcal{D}_\tau = \{(s_t, a_t, g = s_{t+h}, h) : t, h > 0, t+h \leq T \}, \label{eq:dataset}
\end{equation}
with states $s_t$, corresponding actions $a_t$, the corresponding goal set to future state $s_{t+h}$ and matching horizon $h$. Because the relabeling procedure is valid for any horizon, we can use any valid combination of $(s_t, a_t, s_{t+h},h)$ tuples as supervision, for a total of ${T \choose 2}$ optimal datapoints of $(s, a, g, h)$ from a single trajectory. This idea is related to data-relabeling for estimating the value function~\citep{kaelbling1993goals, andrychowicz2017her, rauber2017hindsight}, but our work shows that data-relabelling can also be used to re-interpret data from a sub-optimal agent as though the data came from an optimal agent (with a different goal). 

We then use this relabeled dataset for goal-conditioned behavior cloning. Algorithm~\ref{gcsl-algo} summarizes the approach: 
(1)~Sample a goal from a target goal distribution $p(g)$. (2)~Execute the current policy $\pi(a|s, g, h)$ for $T$ steps in the environment to collect a potentially suboptimal trajectory $\tau$. (3)~Relabel the trajectory (Equation.~\ref{eq:dataset}) to add ${T \choose 2}$ new expert tuples $(s_{t}, a_{t}, s_{t+h}, h)$ to the training dataset. (4)~Perform supervised learning on the entire dataset to update the policy $\pi(a|s, g, h)$ via maximum likelihood. We term this iterative procedure of sampling trajectories, relabeling them, and training a policy until convergence \textit{goal-conditioned supervised learning} (GCSL). This algorithm can use all of the prior off-policy data in the training dataset because this data continues to remain optimal under the notion of goal-reaching optimality that was defined in Section~\ref{sec:background}, but does not require any explicit value function learning.Perhaps surprisingly, this procedure optimizes a lower bound on an RL objective, as we will show in Section~\ref{subsec:theory}.

The GCSL algorithm (as described above) can learn to reach goals from the target distribution $p(g)$ simply using iterated behavioral cloning. This goal reaching algorithm is off-policy, optimizes a simple supervised learning objective, and is easy to implement and tune without the need for any explicit reward function engineering or demonstrations. Additionally, since GCSL uses a goal-conditioned imitation learning algorithm as a sub procedure, if demonstrations or off-policy data are available, it is easier to incorporate this data into training than with off-policy value function methods. 

\begin{algorithm}[t]
  \caption{Goal-Conditioned Supervised Learning~(GCSL)}\label{gcsl-algo}
  \begin{algorithmic}[1]
    \State Initialize policy $\pi_1(\cdot \mid s, g, h)$
    \State Initialize dataset $\mathcal{D}((s,a,g,h))$
    \For{$k= 1, 2, 3, \dots $}
      \State Sample $g \sim p(g)$, collect data with $\pi_k(\cdot \mid \cdot, g)$.
      \State Log trajectory $\tau = (s_0, a_0, s_1, a_1, \dots s_T,a_T)$
      \State Add tuples $\gD_\tau$ to dataset $\gD$ \Comment{see Eq.~\ref{eq:dataset}}
        \State $\pi_{k+1} \gets \argmax_{\pi_\theta} \mathbb{E}_{\mathcal{D}} \left[ \log \pi_{\theta}(a \mid s, g, h) \right]$
    \EndFor
  \end{algorithmic}
\end{algorithm}

\subsection{Theoretical Analysis}
\label{subsec:theory}

We now formally analyze GCSL to verify that it solves the goal-reaching problem, quantify how errors in approximation of the objective manifest in goal-reaching performance, and understand how it relates to existing RL algorithms. Specifically, we derive the algorithm as the optimization of a lower bound of the true goal-reaching objective, and we show that under certain conditions on the environment, minimizing the GCSL objective enables performance guarantees on the learned policy.

We start by describing the objective function being optimized by GCSL. For ease of presentation, we make the simplifying assumption that the trajectories are collected from a single policy $\oldpi$, and that relabelling is only done with goals at the last timestep ($g=s_T$). GCSL performs goal-conditioned behavioral cloning on a distribution of trajectories $\oldpi(\tau) = \mathbb{E}_{g \sim p(g)}[\oldpi(\tau | g)]$, resulting in the following objective:
\vspace{-1em}
\begin{align*}
    J_{\text{GCSL}}(\pi) &= \E_{\tau \sim \oldpi(\tau)}\left[\sum_{t=0}^T \log \pi(a=a_t|s=s_t, g=s_T, h=T-t)\right].
\end{align*}
Our main result shows that, under certain assumptions about the off-policy data distribution, optimizing the GCSL objective $J_{\text{GCSL}}(\pi)$ optimizes a lower bound on the desired objective, $J(\pi)$. %
\begin{theorem}
\label{thm:main_theorem}
Let $J_{\text{GCSL}}$ and $J$ be as defined above. Then, \[J(\pi) \ge J_{\text{GCSL}}(\pi) - 4T(T-1)\alpha^2 + C.\]
Where $\alpha = \max_{s, g, h} D_{TV}(\pi(\cdot|s, g,h) \| \oldpi(\cdot | s, g, h))$ and C is a constant independent of $\pi$.
\end{theorem}
The proof is in Appendix~\ref{appendix:proof}.
This theorem provides a lower-bound on the goal-reaching objective with equality for the optimal policy; akin to many proofs for direct policy search methods, the strongest guarantees are provided under on-policy data collection ($\alpha =0$). The analysis raises two questions: can we quantify the tightness of the bound given by Theorem \ref{thm:main_theorem}, and what does an optimal solution to the GCSL objective imply about performance on the true objective?

The tightness of the bound depends on two choices in the algorithm: how
off-policy data from $\oldpi$ is used to optimize the objective, and how the relabeling step adjusts the exact distribution of data being trained on. 
We find that the looseness induced by the relabeling can be controlled by two factors: 1) the proportion of data that must be relabeled, and 2) the distance between the distribution of trajectories that needed to be relabeled and the distribution of trajectories that achieved the desired goal and were not relabeled. If either of these quantities is minimized to zero, the looseness of the bound that stems from relabeling also goes to zero. We present this analysis formally in Appendix~\ref{appendix:proof-analysis}. 

Even when data is collected from an off-policy distribution, optimizing the GCSL objective over the full state space can provide guarantees on the performance of the learned policy. We write $\pi^*$ to denote a policy that maximizes the true performance $J(\pi)$, and $\Tilde{\pi}^*$ to denote the policy that maximizes the GCSL objective $J_{\text{GCSL}}(\pi)$ over the set of all policies. The following theorem provides such a performance guarantee for deterministic environments (proof in Appendix~\ref{appendix:proof-performance}):
\begin{theorem}
\label{thm:performance_main}
Consider an environment with deterministic dynamics and a data-collection policy $\oldpi$ with full support. If $\max_{s, g, h} D_{TV}(\pi(a | s, g, h), \Tilde{\pi}^*(a | s, g, h)) \leq \epsilon$, then $J(\pi^*) - J(\pi) < \epsilon T$.
\end{theorem}
This theorem states that in an environment with deterministic transitions, the policy that maximizes the GCSL objective $J_{GCSL}(\pi)$ also maximizes the true performance $J(\pi)$.
Furthermore, if the GCSL loss is approximately 
minimized, then performance guarantees can be given as a function of the error across the full state space. Whereas Theorem \ref{thm:main_theorem} shows that GCSL \emph{always} optimizes a lower bound on the RL objective when iteratively re-collecting data with the updated policy, Theorem \ref{thm:performance_main} shows that in certain environments, simply optimizing the GCSL objective from any off-policy data distribution without iterative data collection can also lead to convergence.%

\section{Related Work}
\label{sec:related}

Our work studies the problem of goal-conditioned RL ~\citep{kaelbling1993goals} from sparse goal-reaching rewards. To maximize data-efficiency in the presence of sparse rewards, value function methods use off-policy hindsight relabeling methods such as hindsight experience replay~\citep{andrychowicz2017her} to relabel rewards and transitions retroactively~\citep{schaul2015universal, pong2018temporal}. Despite the potential for learning with hindsight, optimization of goal-conditioned value functions suffers from instability due to challenging critic estimation. \citet{rauber2017hindsight} extends hindsight relabelling to policy gradient methods, but is hampered by high-variance importance weights that emerge from relabelling. Our method also relabels trajectories in hindsight, but does so in a completely different way: to supervise an imitation learning primitive to learn the optimal policy. Unlike these methods, GCSL does not maintain or estimate a value function, enabling a more stable learning problem, and more easily allowing the algorithm to incorporate off-policy data. 

GCSL is inspired by supervised imitation learning~\citep{billard2008robot, hussein2017imitation} via behavioral cloning~\citep{pomerleau1989alvinn}. Recent works have also considered imitation learning with goal relabeling for learning from human play data \citep{lynch2019learning, gupta2019relay} or demonstrations~\citep{ding2019goal}. While GCSL is procedurally similar to \citet{lynch2019learning} and \citet{ding2019goal}, it differs crucially on the type of data used to train the policy --- GCSL is trained on data collected by the agent itself from \emph{scratch}, not from an expert or (noisy) optimal supervisor. The fact that the same algorithmic procedure for training on optimal demonstrations can be applied iteratively using data from a sub-optimal agent to learn from scratch is non-trivial and constitutes one of our contributions. %

GCSL has strong connections to direct policy search and self-imitation algorithms. Direct policy search methods \citep{mannor2003cem, peters2007rwr, theodorou2010pi2, goschin2013cemquantiles, norouzi2016raml, nachum2016urex} selectively weight policies or trajectories by their performance during learning, as measured by the environment's reward function or a learned value function, and maximize the likelihood of these trajectories using supervised learning. Similar algorithmic procedures have also been studied in the context of learning models for planning~\citep{pathak2018zero, savinov2018semi, eysenbach2019search}. GCSL is also closely related to self-imitation learning, where a small subset of trajectories are chosen to be imitated alongside an RL objective \citep{oh2018sil, Hao2019IndependentGA}, often measured using a well-shaped reward function. However, GCSL neither relies on a hand-shaped reward function nor chooses a select group of elites, instead using goal relabeling to imitate every previously collected trajectory for higher data re-use and sample efficiency.

\section{Experimental Evaluation}
In our experiments, we comparatively evaluate GCSL on a number of goal-conditioned tasks. We focus on answering the following questions:
\begin{enumerate}[topsep=0mm,itemsep=0.1mm]
    \item Does GCSL effectively learn goal-conditioned policies from scratch?
    \item Can GCSL learn behaviors more effectively than standard RL methods?
    \item Is GCSL less sensitive to hyperparameters than value-based methods?
    \item Can GCSL incorporate demonstration data more effectively than value-based methods?
\end{enumerate}

\subsection{Experimental Framework}
\label{subsec:exp_details}

\begin{figure}[t]
    \centering
    \begin{subfigure}[t]{0.15\columnwidth}
        \centering
        \includegraphics[width=\linewidth]{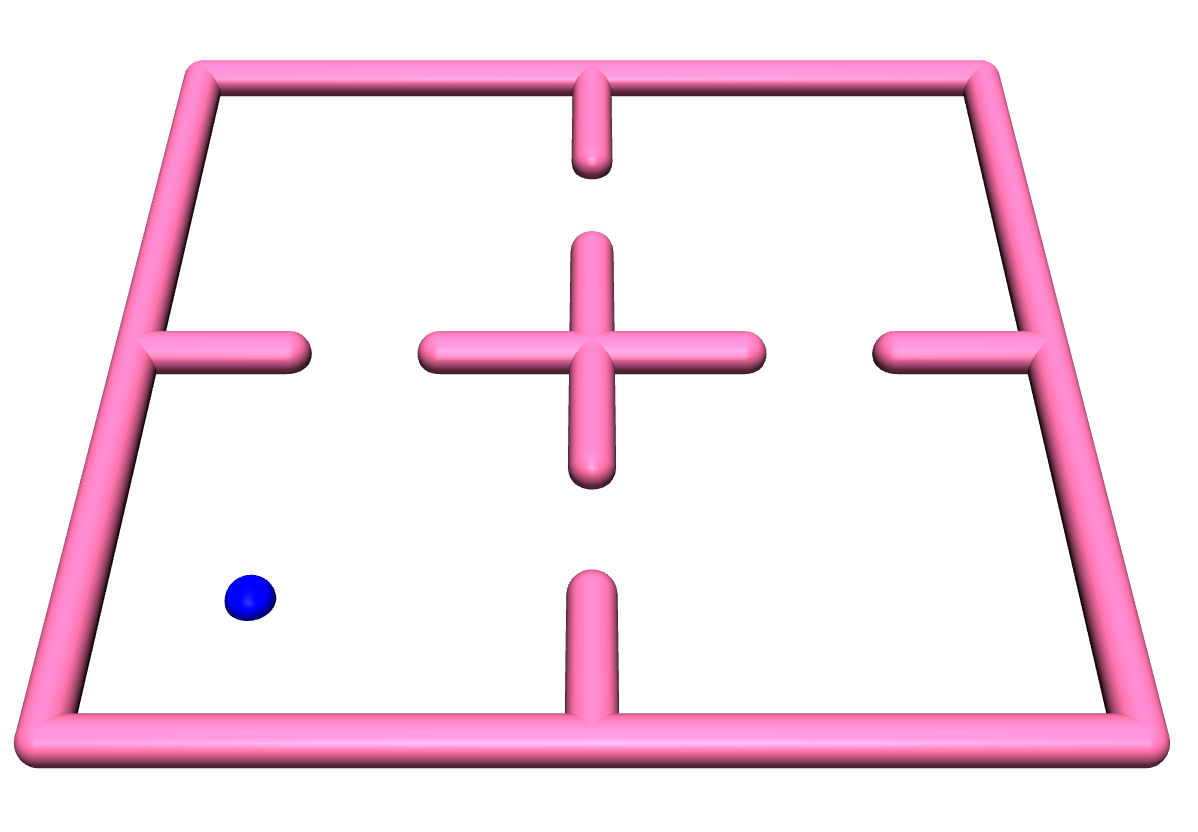}
    \end{subfigure}
    \begin{subfigure}[t]{0.15\columnwidth}
        \centering
        \includegraphics[width=0.8\linewidth]{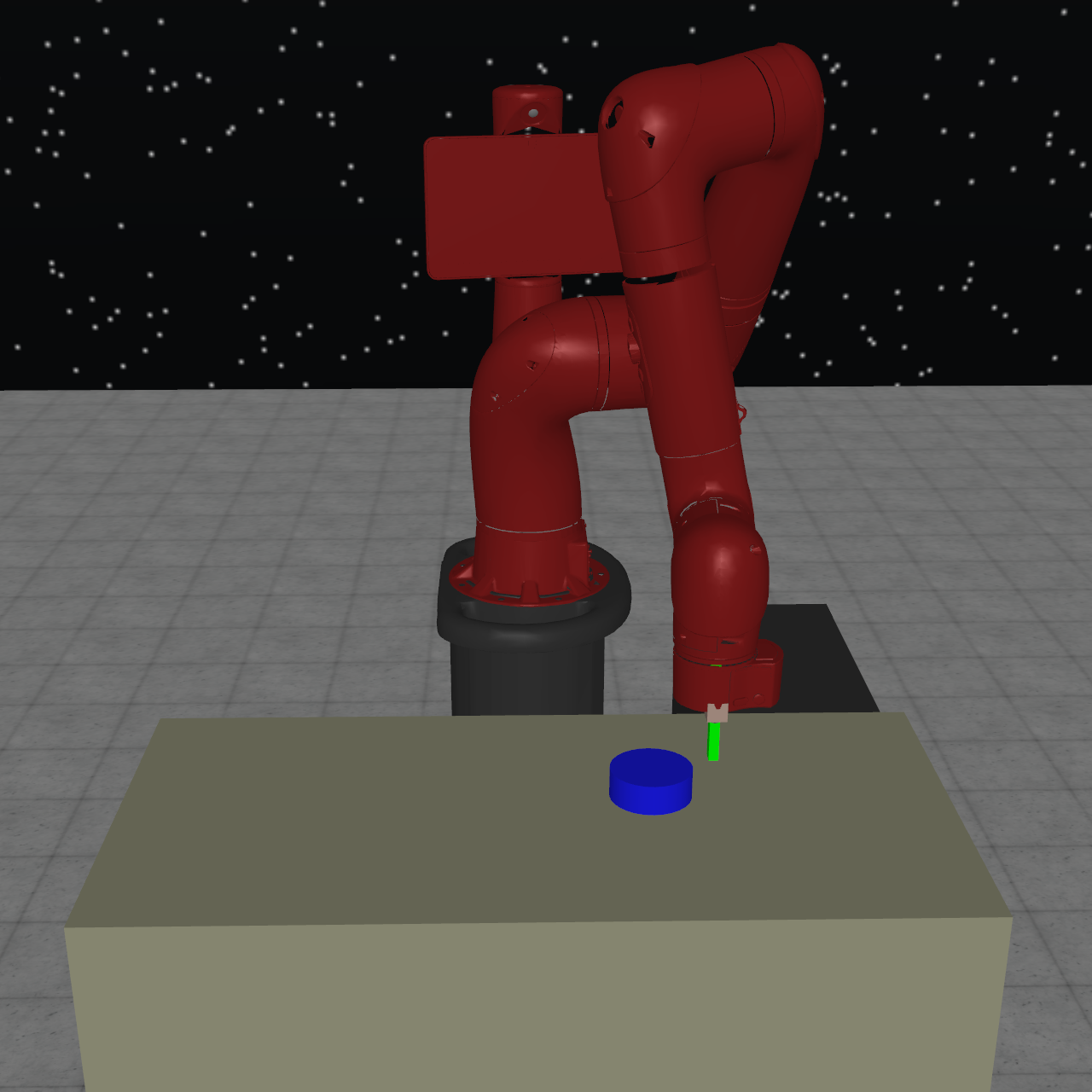}
    \end{subfigure}
    \begin{subfigure}[t]{0.15\columnwidth}
        \centering
        \includegraphics[width=0.8\linewidth]{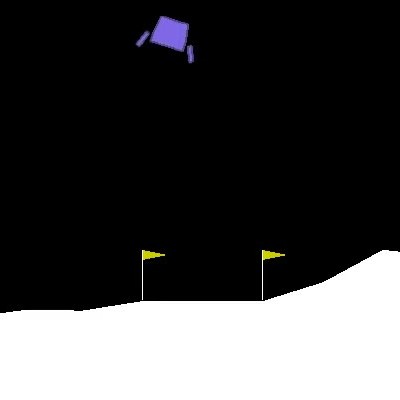}
    \end{subfigure}
    \begin{subfigure}[t]{0.15\columnwidth}
        \centering
        \includegraphics[width=0.8\linewidth]{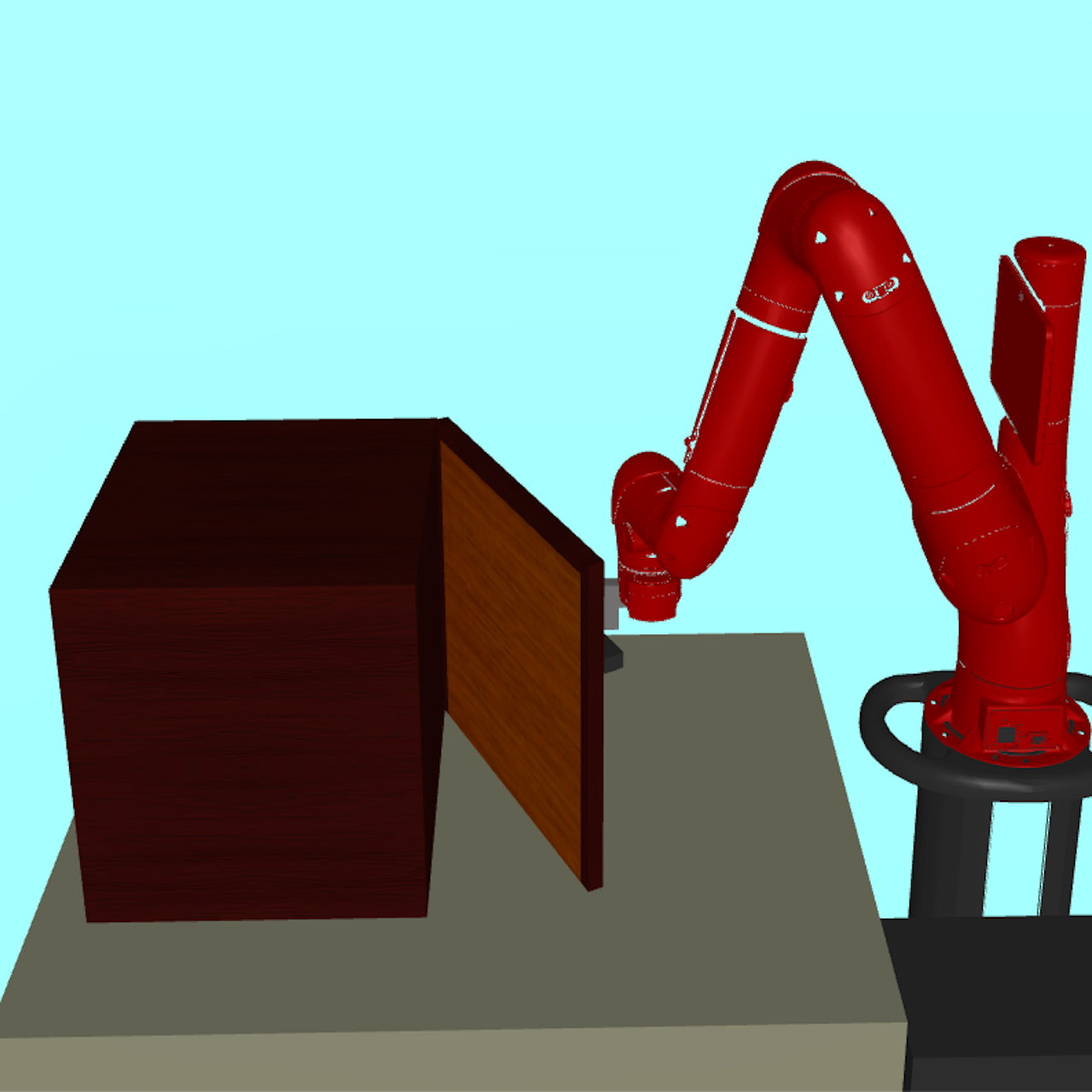}
    \end{subfigure}
    \begin{subfigure}[t]{0.15\columnwidth}
        \centering
        \includegraphics[width=0.8\linewidth]{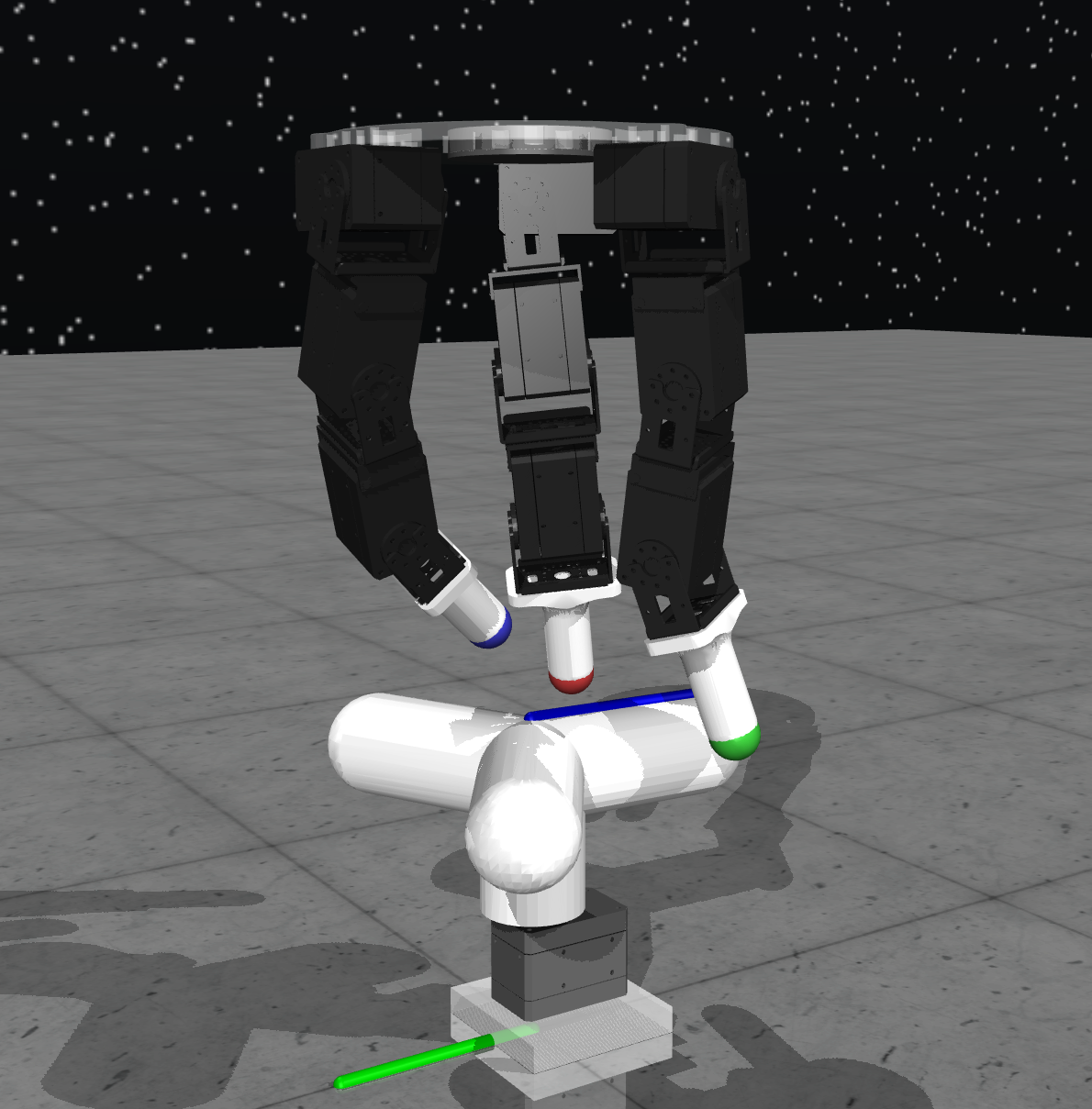}
    \end{subfigure}
     \caption{\textbf{Evaluation Tasks}: We study the following goal-reaching tasks: (from left to right) \\ 2D navigation, robotic pushing, Lunar Lander, robotic door opening, dexterous object manipulation.}
    \label{fig:taskviz}
    \vspace{-1em}
\end{figure}

We evaluate GCSL on five simulated control environments for goal-reaching: 2D room navigation, object pushing with a robotic arm, the classic Lunar Lander game, opening a door with a robotic arm, and object manipulation with a dexterous 9 DoF robotic hand (referred to as claw manipulation), shown in Figure~\ref{fig:taskviz} \citep[Environments from][details in Appendix \ref{appendix:environment-details}]{nair2018rig, ghosh2019learning, ahn2019robel}. These tasks allow us to study the performance of our method under a variety of system dynamics, in settings with both easy and difficult exploration. For each task, the target goal distribution corresponds to a uniform distribution over reachable configurations. Performance is quantified by the distance of the agent to the goal at the last timestep. We present details about the environments, evaluation protocol, hyperparameters, and an extended set of results in Appendix \ref{appendix:experiment-details}.

For the practical implementation of GCSL, we parameterize the policy as a neural network that takes in state, goal, and horizon as input, and outputs a distribution over actions. We found that GCSL performs well even when the horizon is not provided to the policy, despite the optimal policy likely being non-Markovian.
Implementation details for GCSL are in Appendix \ref{appendix:gcsl-details}.

\subsection{Learning Goal-Conditioned Policies}

We first evaluate the effectiveness of GCSL for reaching goals on the domains visualized in Figure \ref{fig:taskviz}, covering a variety of control problems spanning robotics and video games. To better understand the performance of our algorithm, we provide comparisons to value-based methods utilizing hindsight experience replay (HER)~\citep{andrychowicz2017her}, and policy-gradient methods,
two well established families of RL algorithms for solving goal-conditioned tasks. In particular, we compare against TD3-HER,
an off-policy temporal difference RL algorithm that combines TD3~\citep{fujimoto2018off} (an improvement on the DDPG method used by \citet{andrychowicz2017her}) with HER. TD3-HER requires significantly more machinery than GCSL: while GCSL only maintains a policy, TD3-HER maintains a policy, a value function, a target policy, and a target value function, all of which are necessary for good performance. We also compare with PPO~\citep{schulman2017ppo}, a state-of-the-art on-policy policy gradient algorithm that does not leverage data relabeling, but is known to provide more stable optimization than off-policy methods and perform well on typical benchmark problems. Details for the training procedure for these comparisons, hyperparameter and architectural choices, as well as some additional comparisons are presented in Appendix~\ref{appendix:rl-details}. 

\begin{figure*}[t]
    \centering
    \vspace{-3em}
    \includegraphics[width=0.9\linewidth]{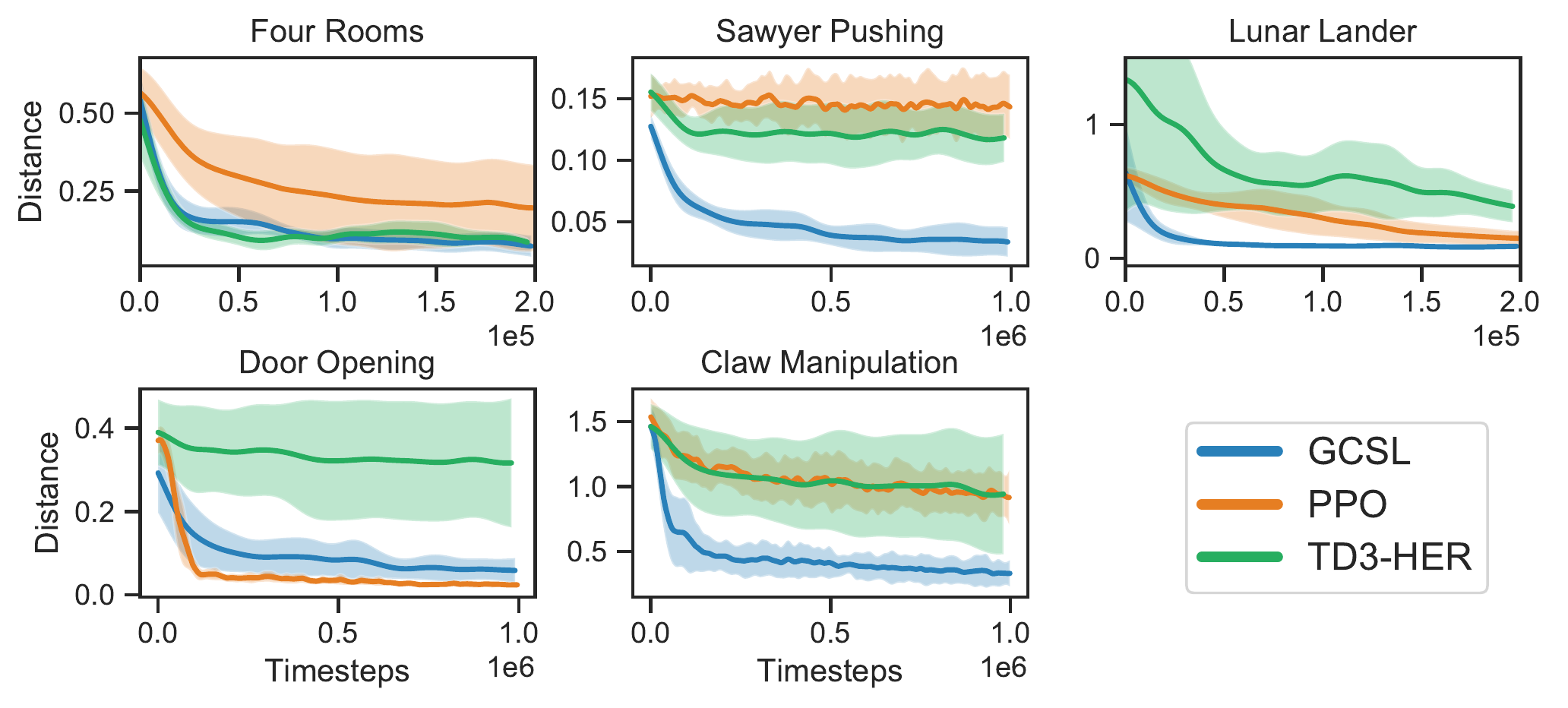}
    \vspace{-1em}
    \caption{On a majority of tasks, GCSL performs well or better compared to more complex RL algorithms like PPO~\citep{schulman2017ppo} or TD3-HER~\citep{andrychowicz2017her}. Shaded regions denote the standard deviation across {5} random seeds (lower is better). }
    \label{fig:from-state}
    \vspace{-1.5em}
\end{figure*}

The results in Figure \ref{fig:from-state} show that GCSL generally performs as well or better than the best performing prior RL method on each task, only losing out slightly to PPO on the door opening task, where exploration is less of a challenge. GCSL outperforms both methods by a large margin on the pushing and claw tasks, and by a small margin on the lunar lander task. These empirical results suggest that GCSL, despite its simplicity, represents a stable and appealing alternative to significantly more complex RL methods, without the need for separate critics, policy gradients, or target networks. 

\subsection{Analysis of Learning Progress and Learned Behaviors}
\label{subsection:ablations}

\begin{wrapfigure}{R}{0.4\textwidth}
\centering
    \vspace{-2em}
    \includegraphics[width=\linewidth]{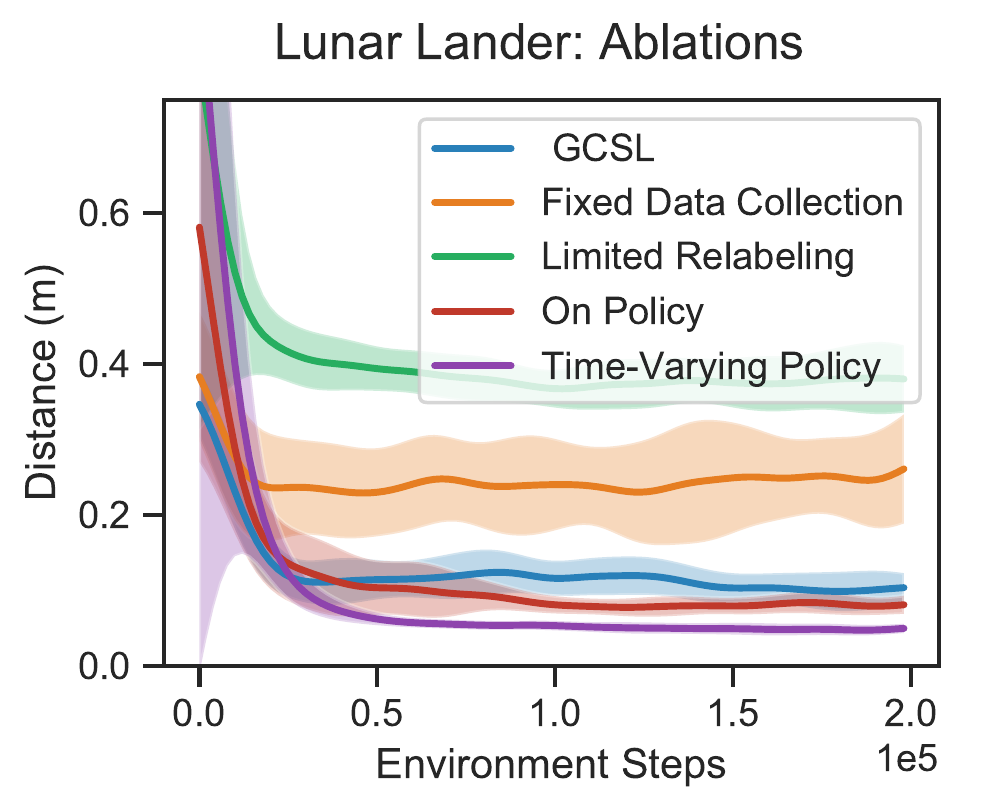}
    \includegraphics[width=\linewidth]{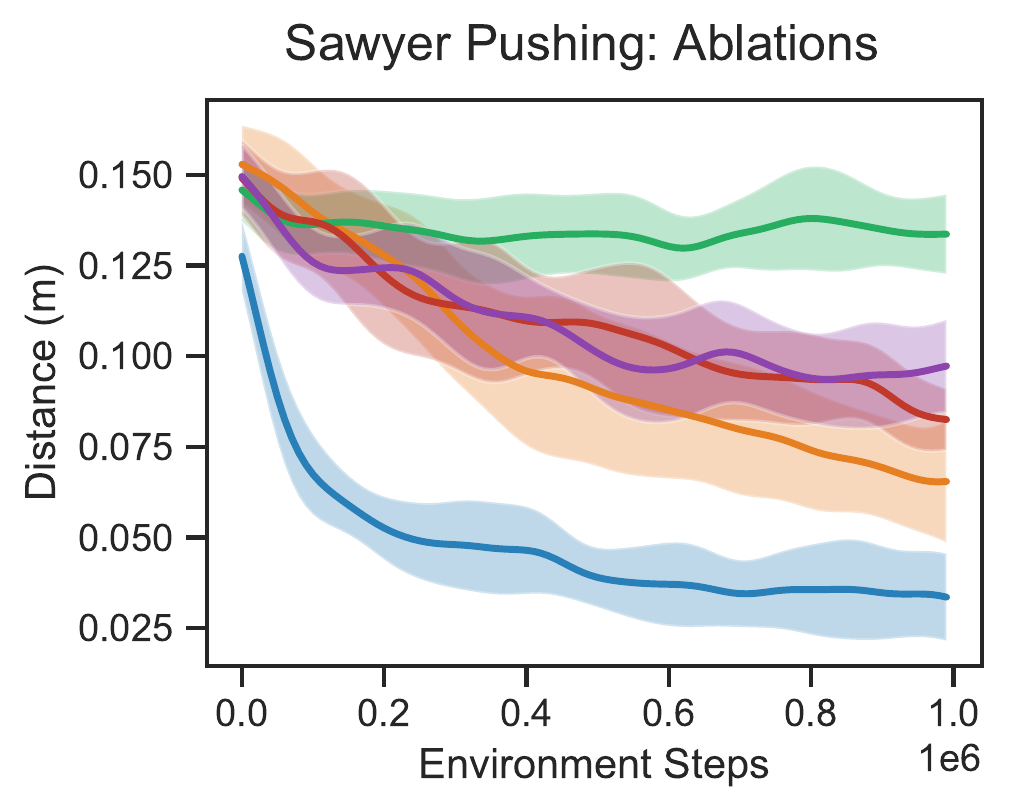}
    \caption{{ {Ablations} of GCSL on the Lunar Lander and robotic pushing domains. Other domains in Appendix \ref{appendix:ablation-details}.}}
    \vspace{-1em}
     \label{fig:ablations}
\end{wrapfigure}
To analyze GCSL, we evaluate its performance in a number of scenarios, varying the quality and quantity of data, the policy class, and the relabeling technique (Figure \ref{fig:ablations}). Full details for these scenarios and results for all domains are in Appendix \ref{appendix:ablation-details}. 

First, we study how varying the policy class can affect the performance of GCSL. In Section \ref{subsec:exp_details}, we hypothesized that GCSL with a Markovian policy would outperform a time-varying policy. Indeed, allowing policies to be time-varying (``Time-Varying Policy" in Figure \ref{fig:ablations}) speeds up training on domains like Lunar Lander; on domains requiring more exploration like the Sawyer pushing task, exploration using time-varying policies is ineffective and degrades performance.

To investigate the impact of the data-collection policy, we consider variations that collect data using a fixed policy or train only on on-policy data. When collecting data using a fixed policy (``Fixed Data Collection" in Figure \ref{fig:ablations}), the algorithm learns much slower, suggesting that iterative data collection is crucial for GCSL. By forcing the data to be on-policy (``On-Policy" in Figure \ref{fig:ablations}), the algorithm cannot utilize all data seen during training. GCSL still makes progress in this case, but more slowly. We additionally consider limited-horizon relabeling, in which only states and goals that are at most $3$
steps apart are relabeled, similar to proposals in prior work~\citep{pathak2018zero, savinov2018semi}. Limiting the horizon degrades performance (``Limited relabeling" in Figure \ref{fig:ablations}),
indicating that multi-horizon relabeling is important.

Finally, we discuss the concern that since GCSL uses final-timestep optimality, it may provide significantly different behaviors than shortest-path optimality. While in theory, GCSL can learn round-about trajectories or otherwise exhibit pathological behavior, we find that on our empirical benchmarks, GCSL learns fairly direct goal-reaching behaviors (visualized in Appendix \ref{appendix:example_trajectories}). Since even the time-varying policy shares network parameters for different horizons, we hypothesize that the policy is constrained to produce behaviors that are roughly consistent through time, resulting in directed behaviors that resemble shortest-path optimality.

\subsection{Robustness to Hyperparameters}
\label{sec:robustness}

\begin{figure}[t]
    \centering
    \vspace{-1.5em}
       \includegraphics[width=\linewidth]{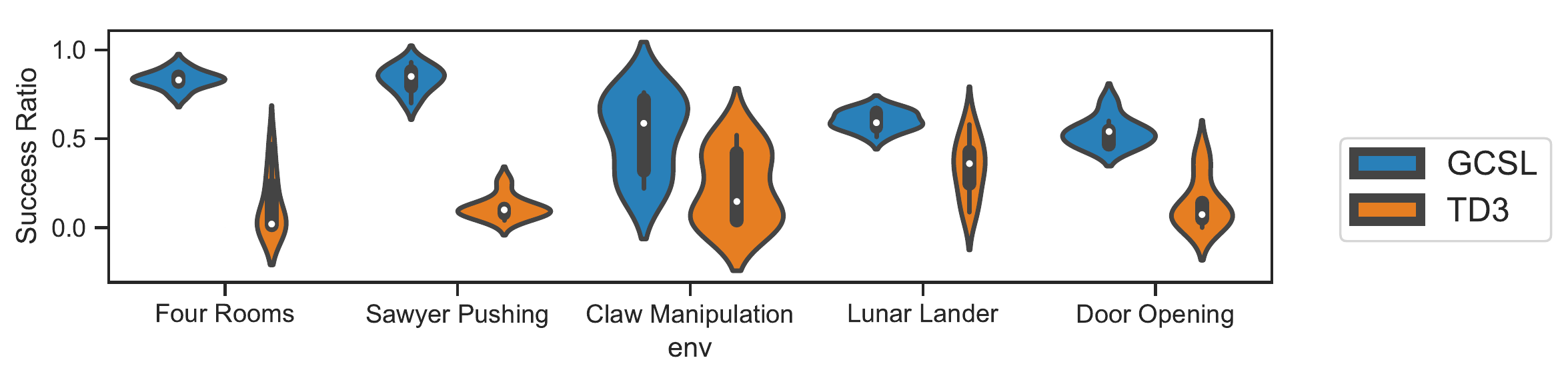}
    \vspace{-1em}
    \caption{\textbf{Hyperparameter Robustness}: Distribution of final performance of GCSL and TD3-HER across nine hyperparameter configurations in each environment (see Section \ref{sec:robustness} for details). Higher values indicate better performance, and tightly clustered distributions indicate lower sensitivity to hyperparameters. GCSL is more performant and robust to hyperparameters than TD3-HER.}
    \label{fig:robustness}
    \vspace{-1em}
\end{figure}

Our next experiment tests the hypothesis that GCSL is more robust to hyperparameters than value-based RL methods like TD3-HER. The intuition is that, while dynamic programming methods are known to be quite sensitive to hyperparameters~\citep{henderson2018deep}, supervised learning techniques seem more robust. We ran a sweep across nine hyperparameter configurations, varying network capacity (size of the hidden layers in $[250, 500, 1000]$) and frequency of gradient updates (gradient updates per environment step in $[1, 2, 4]$). We compared both GCSL and TD3-HER and plotted the distribution of final timestep performance across all possible configurations in Fig.~\ref{fig:robustness}. We observe that the distribution of performance for GCSL is more tightly clustered than for TD3-HER, indicating lower sensitivity to hyperparameters. We emphasize that GCSL has fewer hyperparameters than TD3-HER; since GCSL does not learn a value function, it does not require parameters for the value function architecture, target update frequency, discount factor, or actor update frequency.

\subsection{Initializing with Demonstrations}
\label{sec:demo-experiment}
\begin{wrapfigure}{R}{0.4\textwidth}
\centering
    \includegraphics[width=\linewidth]{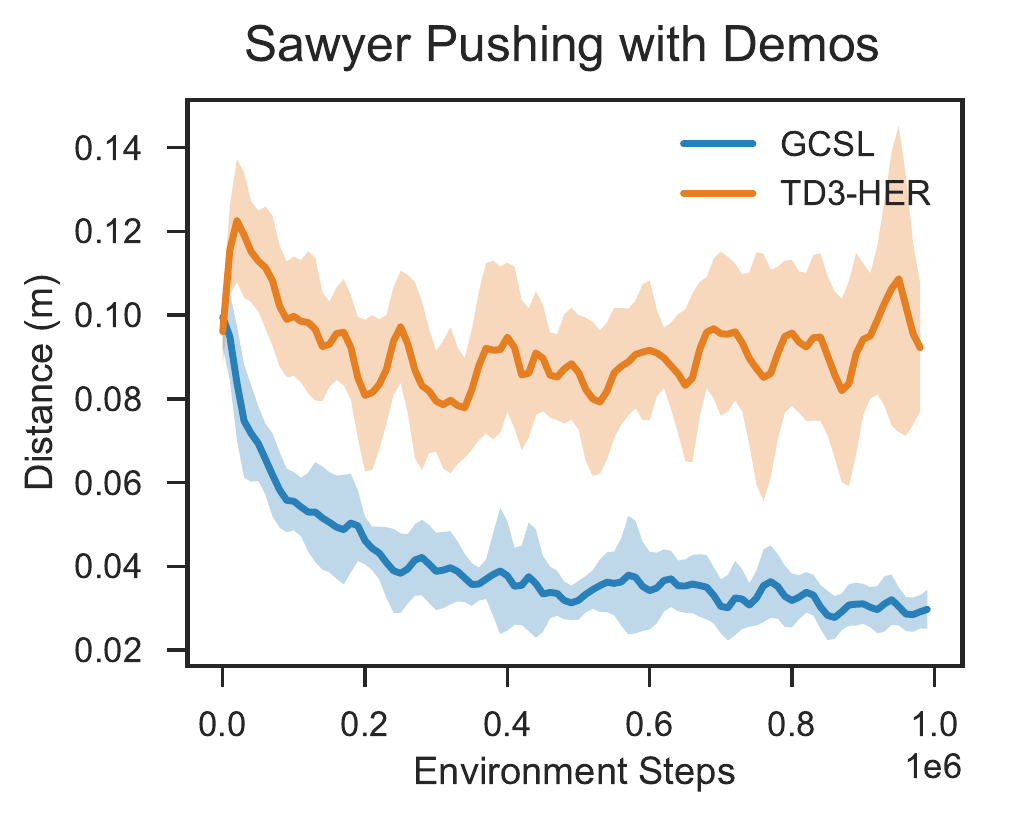}
    \caption{{\textbf{Demonstrations:} GCSL incorporates expert demonstrations more effectively than TD3-HER.}}
\label{fig:demobootstrap}
\end{wrapfigure}
As GCSL can relabel and imitate trajectories from arbitrary sources, the algorithm is amenable to initialization from logs of previously collected trajectories or from demonstration data collected by an expert. In this section, we compare the performance of GCSL bootstrapped from expert demonstrations to TD3-HER. Both methods can in principle utilize off-policy demonstrations; however, our results in Figure~\ref{fig:demobootstrap} show that GCSL benefits substantially more from these demonstrations. While value-based RL methods are known to struggle with data that is far off-policy~\citep{kumar2019bear}, the simple supervised learning procedure in GCSL can take advantage of such data easily.

In this experiment, we provide the agent with a set of demonstration trajectories, each for reaching a different goal. GCSL adds this data to the initial dataset, without any other modifications to the algorithm.
For TD3-HER, we incorporate demonstrations following the setup of \citet{vecerik2017leveraging}. Even with these measures, the value function in TD3-HER still suffers degraded performance and error accumulation during pre-training. When expert demonstrations are provided for the robotic pushing environment (Figure \ref{fig:demobootstrap}), GCSL progressively improves faster than when from scratch, but TD3 is unable to improve substantially beyond the original behavioral-cloned policy. We hypothesize that the difference in performance largely occurs because of the instability and optimism bias present when training value functions using~demonstrations.

\section{Discussion and Future Work}
We proposed GCSL, a simple algorithm that uses supervised learning on its own previously collected data to iteratively learn goal-reaching policies from scratch. GCSL lifts several limitations of previous goal-reaching methods: it does not require a hand-defined reward, expert demonstrations, or the need to learn a value function. GCSL often outperforms more complex RL algorithms, is robust to hyperparameters, uses off-policy data, and can incorporate expert demonstrations when they are available. The current instantiation of GCSL is limited in exploration, since it relies primarily on the stochasticity of the policy to explore; a promising future direction would be to selectively reweight the sampled rollouts to promote novelty-seeking exploration. Nonetheless, GCSL is simple, scalable, and readily applicable --- a step towards the fully autonomous learning of goal-directed agents.

\paragraph{Acknowledgements.} This research was supported by an NSF graduate fellowship, Berkeley DeepDrive, the National Science Foundation, the Office of Naval Research, and support from Google, Amazon, and NVIDIA. We thank Karol Hausman, Ignasi Clavera, Aviral Kumar, Marvin Zhang, Rishabh Agarwal, Daniel D. Johnson, and Vikash Kumar for thoughtful discussions, insights and feedback on paper drafts. 

\bibliography{iclr2021_conference}

\begin{thebibliography}{41}
\providecommand{\natexlab}[1]{#1}
\providecommand{\url}[1]{\texttt{#1}}
\expandafter\ifx\csname urlstyle\endcsname\relax
  \providecommand{\doi}[1]{doi: #1}\else
  \providecommand{\doi}{doi: \begingroup \urlstyle{rm}\Url}\fi

\bibitem[Henderson et~al.(2018)Henderson, Islam, Bachman, Pineau, Precup, and
  Meger]{henderson2018deep}
Peter Henderson, Riashat Islam, Philip Bachman, Joelle Pineau, Doina Precup,
  and David Meger.
\newblock Deep reinforcement learning that matters.
\newblock In \emph{Thirty-Second AAAI Conference on Artificial Intelligence},
  2018.

\bibitem[Tsitsiklis and Van~Roy(1997)]{tsitsiklis1997analysis}
John~N Tsitsiklis and Benjamin Van~Roy.
\newblock Analysis of temporal-diffference learning with function
  approximation.
\newblock In \emph{Advances in neural information processing systems}, pages
  1075--1081, 1997.

\bibitem[van Hasselt et~al.(2018)van Hasselt, Doron, Strub, Hessel, Sonnerat,
  and Modayil]{Hasselt2018DeepRL}
Hado van Hasselt, Yotam Doron, Florian Strub, Matteo Hessel, Nicolas Sonnerat,
  and Joseph Modayil.
\newblock Deep reinforcement learning and the deadly triad.
\newblock \emph{ArXiv}, abs/1812.02648, 2018.

\bibitem[Kumar et~al.(2019{\natexlab{a}})Kumar, Fu, Tucker, and
  Levine]{kumar2019stabilizing}
Aviral Kumar, Justin Fu, George Tucker, and Sergey Levine.
\newblock Stabilizing off-policy q-learning via bootstrapping error reduction.
\newblock \emph{arXiv preprint arXiv:1906.00949}, 2019{\natexlab{a}}.

\bibitem[Bojarski et~al.(2016)Bojarski, Testa, Dworakowski, Firner, Flepp,
  Goyal, Jackel, Monfort, Muller, Zhang, Zhang, Zhao, and
  Zieba]{bojarski2016imitation}
Mariusz Bojarski, Davide~Del Testa, Daniel Dworakowski, Bernhard Firner, Beat
  Flepp, Prasoon Goyal, Lawrence~D. Jackel, Mathew Monfort, Urs Muller, Jiakai
  Zhang, Xin Zhang, Jake Zhao, and Karol Zieba.
\newblock End to end learning for self-driving cars.
\newblock \emph{CoRR}, abs/1604.07316, 2016.
\newblock URL \url{http://arxiv.org/abs/1604.07316}.

\bibitem[Lynch et~al.(2019)Lynch, Khansari, Xiao, Kumar, Tompson, Levine, and
  Sermanet]{lynch2019learning}
Corey Lynch, Mohi Khansari, Ted Xiao, Vikash Kumar, Jonathan Tompson, Sergey
  Levine, and Pierre Sermanet.
\newblock Learning latent plans from play.
\newblock \emph{arXiv preprint arXiv:1903.01973}, 2019.

\bibitem[Kaelbling(1993)]{kaelbling1993goals}
Leslie~Pack Kaelbling.
\newblock Learning to achieve goals.
\newblock In \emph{International Joint Conference on Artificial Intelligence
  (IJCAI)}, pages 1094--1098, 1993.

\bibitem[Andrychowicz et~al.(2017)Andrychowicz, Wolski, Ray, Schneider, Fong,
  Welinder, McGrew, Tobin, Abbeel, and Zaremba]{andrychowicz2017her}
Marcin Andrychowicz, Filip Wolski, Alex Ray, Jonas Schneider, Rachel Fong,
  Peter Welinder, Bob McGrew, Josh Tobin, OpenAI~Pieter Abbeel, and Wojciech
  Zaremba.
\newblock Hindsight experience replay.
\newblock In \emph{Advances in Neural Information Processing Systems}, pages
  5048--5058, 2017.

\bibitem[Rauber et~al.(2017)Rauber, Ummadisingu, Mutz, and
  Schmidhuber]{rauber2017hindsight}
Paulo Rauber, Avinash Ummadisingu, Filipe Mutz, and Juergen Schmidhuber.
\newblock Hindsight policy gradients.
\newblock \emph{arXiv preprint arXiv:1711.06006}, 2017.

\bibitem[Gupta et~al.(2019)Gupta, Kumar, Lynch, Levine, and
  Hausman]{gupta2019relay}
Abhishek Gupta, Vikash Kumar, Corey Lynch, Sergey Levine, and Karol Hausman.
\newblock Relay policy learning: Solving long-horizon tasks via imitation and
  reinforcement learning.
\newblock \emph{CoRR}, abs/1910.11956, 2019.
\newblock URL \url{http://arxiv.org/abs/1910.11956}.

\bibitem[Ding et~al.(2019)Ding, Florensa, Phielipp, and Abbeel]{ding2019goal}
Yiming Ding, Carlos Florensa, Mariano Phielipp, and Pieter Abbeel.
\newblock Goal conditioned imitation learning.
\newblock In \emph{Advances in Neural Information Processing Systems}, 2019.

\bibitem[Oh et~al.(2018)Oh, Guo, Singh, and Lee]{oh2018sil}
Junhyuk Oh, Yijie Guo, Satinder Singh, and Honglak Lee.
\newblock Self-imitation learning.
\newblock In \emph{International Conference on Machine Learning}, pages
  3875--3884, 2018.

\bibitem[Hao et~al.(2019)Hao, Wang, Hao, and Yang]{Hao2019IndependentGA}
Xiaotian Hao, Weixun Wang, Jianye Hao, and Y.~Yang.
\newblock Independent generative adversarial self-imitation learning in
  cooperative multiagent systems.
\newblock In \emph{AAMAS}, 2019.

\bibitem[Neumann and Peters(2009)]{neumann2009fitted}
Gerhard Neumann and Jan~R Peters.
\newblock Fitted q-iteration by advantage weighted regression.
\newblock In \emph{Advances in neural information processing systems}, pages
  1177--1184, 2009.

\bibitem[Abdolmaleki et~al.(2018)Abdolmaleki, Springenberg, Tassa, Munos,
  Heess, and Riedmiller]{abdolmaleki2018maximum}
Abbas Abdolmaleki, Jost~Tobias Springenberg, Yuval Tassa, Remi Munos, Nicolas
  Heess, and Martin Riedmiller.
\newblock Maximum a posteriori policy optimisation.
\newblock \emph{arXiv preprint arXiv:1806.06920}, 2018.

\bibitem[Peng et~al.(2019)Peng, Kumar, Zhang, and Levine]{peng2019advantage}
Xue~Bin Peng, Aviral Kumar, Grace Zhang, and Sergey Levine.
\newblock Advantage-weighted regression: Simple and scalable off-policy
  reinforcement learning.
\newblock \emph{arXiv preprint arXiv:1910.00177}, 2019.

\bibitem[Schaul et~al.(2015)Schaul, Horgan, Gregor, and
  Silver]{schaul2015universal}
Tom Schaul, Daniel Horgan, Karol Gregor, and David Silver.
\newblock Universal value function approximators.
\newblock In \emph{International conference on machine learning}, pages
  1312--1320, 2015.

\bibitem[Pong et~al.(2018)Pong, Gu, Dalal, and Levine]{pong2018temporal}
Vitchyr Pong, Shixiang Gu, Murtaza Dalal, and Sergey Levine.
\newblock Temporal difference models: Model-free deep rl for model-based
  control.
\newblock \emph{arXiv preprint arXiv:1802.09081}, 2018.

\bibitem[Billard et~al.(2008)Billard, Calinon, Dillmann, and
  Schaal]{billard2008robot}
Aude Billard, Sylvain Calinon, Ruediger Dillmann, and Stefan Schaal.
\newblock Robot programming by demonstration.
\newblock \emph{Springer handbook of robotics}, pages 1371--1394, 2008.

\bibitem[Hussein et~al.(2017)Hussein, Gaber, Elyan, and
  Jayne]{hussein2017imitation}
Ahmed Hussein, Mohamed~Medhat Gaber, Eyad Elyan, and Chrisina Jayne.
\newblock Imitation learning: A survey of learning methods.
\newblock \emph{ACM Computing Surveys (CSUR)}, 50\penalty0 (2):\penalty0 21,
  2017.

\bibitem[Pomerleau(1989)]{pomerleau1989alvinn}
Dean~A Pomerleau.
\newblock Alvinn: An autonomous land vehicle in a neural network.
\newblock In \emph{Advances in neural information processing systems}, pages
  305--313, 1989.

\bibitem[Mannor et~al.(2003)Mannor, Rubinstein, and Gat]{mannor2003cem}
Shie Mannor, Reuven~Y Rubinstein, and Yohai Gat.
\newblock The cross entropy method for fast policy search.
\newblock In \emph{Proceedings of the 20th International Conference on Machine
  Learning (ICML-03)}, pages 512--519, 2003.

\bibitem[Peters and Schaal(2007)]{peters2007rwr}
Jan Peters and Stefan Schaal.
\newblock Reinforcement learning by reward-weighted regression for operational
  space control.
\newblock In \emph{Proceedings of the 24th international conference on Machine
  learning}, pages 745--750. ACM, 2007.

\bibitem[Theodorou et~al.(2010)Theodorou, Buchli, and Schaal]{theodorou2010pi2}
Evangelos Theodorou, Jonas Buchli, and Stefan Schaal.
\newblock A generalized path integral control approach to reinforcement
  learning.
\newblock \emph{journal of machine learning research}, 11\penalty0
  (Nov):\penalty0 3137--3181, 2010.

\bibitem[Goschin et~al.(2013)Goschin, Weinstein, and
  Littman]{goschin2013cemquantiles}
Sergiu Goschin, Ari Weinstein, and Michael Littman.
\newblock The cross-entropy method optimizes for quantiles.
\newblock In \emph{International Conference on Machine Learning}, pages
  1193--1201, 2013.

\bibitem[Norouzi et~al.(2016)Norouzi, Bengio, Jaitly, Schuster, Wu, Schuurmans,
  et~al.]{norouzi2016raml}
Mohammad Norouzi, Samy Bengio, Navdeep Jaitly, Mike Schuster, Yonghui Wu, Dale
  Schuurmans, et~al.
\newblock Reward augmented maximum likelihood for neural structured prediction.
\newblock In \emph{Advances In Neural Information Processing Systems}, pages
  1723--1731, 2016.

\bibitem[Nachum et~al.(2016)Nachum, Norouzi, and Schuurmans]{nachum2016urex}
Ofir Nachum, Mohammad Norouzi, and Dale Schuurmans.
\newblock Improving policy gradient by exploring under-appreciated rewards.
\newblock \emph{arXiv preprint arXiv:1611.09321}, 2016.

\bibitem[Pathak et~al.(2018)Pathak, Mahmoudieh, Luo, Agrawal, Chen, Shentu,
  Shelhamer, Malik, Efros, and Darrell]{pathak2018zero}
Deepak Pathak, Parsa Mahmoudieh, Guanghao Luo, Pulkit Agrawal, Dian Chen, Yide
  Shentu, Evan Shelhamer, Jitendra Malik, Alexei~A Efros, and Trevor Darrell.
\newblock Zero-shot visual imitation.
\newblock In \emph{Proceedings of the IEEE Conference on Computer Vision and
  Pattern Recognition Workshops}, pages 2050--2053, 2018.

\bibitem[Savinov et~al.(2018)Savinov, Dosovitskiy, and Koltun]{savinov2018semi}
Nikolay Savinov, Alexey Dosovitskiy, and Vladlen Koltun.
\newblock Semi-parametric topological memory for navigation.
\newblock \emph{arXiv preprint arXiv:1803.00653}, 2018.

\bibitem[Eysenbach et~al.(2019)Eysenbach, Salakhutdinov, and
  Levine]{eysenbach2019search}
Benjamin Eysenbach, Ruslan Salakhutdinov, and Sergey Levine.
\newblock Search on the replay buffer: Bridging planning and reinforcement
  learning.
\newblock \emph{arXiv preprint arXiv:1906.05253}, 2019.

\bibitem[Nair et~al.(2018)Nair, Pong, Dalal, Bahl, Lin, and
  Levine]{nair2018rig}
Ashvin~V Nair, Vitchyr Pong, Murtaza Dalal, Shikhar Bahl, Steven Lin, and
  Sergey Levine.
\newblock Visual reinforcement learning with imagined goals.
\newblock In \emph{Advances in Neural Information Processing Systems}, pages
  9191--9200, 2018.

\bibitem[Ghosh et~al.(2019)Ghosh, Gupta, and Levine]{ghosh2019learning}
Dibya Ghosh, Abhishek Gupta, and Sergey Levine.
\newblock Learning actionable representations with goal conditioned policies.
\newblock In \emph{International Conference on Learning Representations}, 2019.

\bibitem[Ahn et~al.(2019)Ahn, Zhu, Hartikainen, Ponte, Gupta, Levine, and
  Kumar]{ahn2019robel}
Michael Ahn, Henry Zhu, Kristian Hartikainen, Hugo Ponte, Abhishek Gupta,
  Sergey Levine, and Vikash Kumar.
\newblock Robel: Robotics benchmarks for learning with low-cost robots, 2019.

\bibitem[Fujimoto et~al.(2018)Fujimoto, Meger, and Precup]{fujimoto2018off}
Scott Fujimoto, David Meger, and Doina Precup.
\newblock Off-policy deep reinforcement learning without exploration.
\newblock \emph{arXiv preprint arXiv:1812.02900}, 2018.

\bibitem[Schulman et~al.(2017)Schulman, Wolski, Dhariwal, Radford, and
  Klimov]{schulman2017ppo}
John Schulman, Filip Wolski, Prafulla Dhariwal, Alec Radford, and Oleg Klimov.
\newblock Proximal policy optimization algorithms.
\newblock \emph{CoRR}, abs/1707.06347, 2017.

\bibitem[Kumar et~al.(2019{\natexlab{b}})Kumar, Fu, Tucker, and
  Levine]{kumar2019bear}
Aviral Kumar, Justin Fu, George Tucker, and Sergey Levine.
\newblock Stabilizing off-policy q-learning via bootstrapping error reduction.
\newblock \emph{CoRR}, abs/1906.00949, 2019{\natexlab{b}}.

\bibitem[Vecerik et~al.(2017)Vecerik, Hester, Scholz, Wang, Pietquin, Piot,
  Heess, Roth{\"o}rl, Lampe, and Riedmiller]{vecerik2017leveraging}
Mel Vecerik, Todd Hester, Jonathan Scholz, Fumin Wang, Olivier Pietquin, Bilal
  Piot, Nicolas Heess, Thomas Roth{\"o}rl, Thomas Lampe, and Martin Riedmiller.
\newblock Leveraging demonstrations for deep reinforcement learning on robotics
  problems with sparse rewards.
\newblock \emph{arXiv preprint arXiv:1707.08817}, 2017.

\bibitem[Brockman et~al.(2016)Brockman, Cheung, Pettersson, Schneider,
  Schulman, Tang, and Zaremba]{brockman2016gym}
Greg Brockman, Vicki Cheung, Ludwig Pettersson, Jonas Schneider, John Schulman,
  Jie Tang, and Wojciech Zaremba.
\newblock Openai gym.
\newblock \emph{CoRR}, abs/1606.01540, 2016.
\newblock URL \url{http://arxiv.org/abs/1606.01540}.

\bibitem[Schulman et~al.(2015)Schulman, Levine, Abbeel, Jordan, and
  Moritz]{schulman2015trpo}
John Schulman, Sergey Levine, Pieter Abbeel, Michael Jordan, and Philipp
  Moritz.
\newblock Trust region policy optimization.
\newblock In Francis Bach and David Blei, editors, \emph{Proceedings of the
  32nd International Conference on Machine Learning}, volume~37 of
  \emph{Proceedings of Machine Learning Research}, pages 1889--1897, Lille,
  France, 07--09 Jul 2015. PMLR.

\bibitem[Kakade and Langford(2002)]{Kakade:2002:AOA:645531.656005}
Sham Kakade and John Langford.
\newblock Approximately optimal approximate reinforcement learning.
\newblock In \emph{Proceedings of the Nineteenth International Conference on
  Machine Learning}, ICML '02, pages 267--274, San Francisco, CA, USA, 2002.
  Morgan Kaufmann Publishers Inc.
\newblock ISBN 1-55860-873-7.
\newblock URL \url{http://dl.acm.org/citation.cfm?id=645531.656005}.

\bibitem[Ross et~al.(2011)Ross, Gordon, and Bagnell]{dagger}
Stephane Ross, Geoffrey Gordon, and Drew Bagnell.
\newblock A reduction of imitation learning and structured prediction to
  no-regret online learning.
\newblock In Geoffrey Gordon, David Dunson, and Miroslav Dudík, editors,
  \emph{Proceedings of the Fourteenth International Conference on Artificial
  Intelligence and Statistics}, volume~15 of \emph{Proceedings of Machine
  Learning Research}, pages 627--635, Fort Lauderdale, FL, USA, 11--13 Apr
  2011. PMLR.
\newblock URL \url{http://proceedings.mlr.press/v15/ross11a.html}.

\end{thebibliography}
\bibliographystyle{iclr2021_conference}

\appendix
 \onecolumn

\section{Experimental Details}
\label{appendix:experiment-details}

\subsection{Goal-Conditioned Supervised Learning (GCSL)}
\label{appendix:gcsl-details}

GCSL iteratively performs maximum likelihood estimation using a dataset of relabeled trajectories that have been previously collected by the agent. Here we present details about the policy class, data collection procedure, and other design choices.

We parameterize a time-invariant policy using a neural network which takes as input state and goal (not the horizon), and returns probabilities for a discretized grid of actions of the action space. The neural network concatenates the state and goal together, and passes the concatenated input into a feedforward network with two hidden layers of size $400$ and $300$ respectively, outputting logits for each discretized action. Empirically, we have found GCSL to perform much better with larger choices of neural networks; however, we use this two-layer neural network for fair comparisons to TD3-HER. The GCSL loss is optimized using the Adam optimizer with learning rate $\alpha=5\times 10^{-4}$, with a batch size of $256$, taking one gradient step for every step in the environment. 

When executing in the environment, the first $10000$ environment steps are taken according to uniform random action selection, after which the data-collection policy is the greedy policy: $a = \arg\max_{a}\pi(a|s,g)$. The replay buffer stores trajectories and relabels on the fly, with the size of the buffer subject only to memory constraints. To clarify, instead of explicitly relabeling and storing all ${T \choose 2}$ possible tuples from a trajectory, we instead save the trajectory and relabel at training time. When sampling from the dataset, a trajectory is chosen at random, a start index $t$ and goal index $t' > t$ are sampled uniformly at random, and the tuple corresponding to this state and goal are relabelled and sampled.

\subsection{RL Comparisons}
\label{appendix:rl-details}

\begin{figure*}[!ht]
    \centering
    \includegraphics[width=0.8\linewidth]{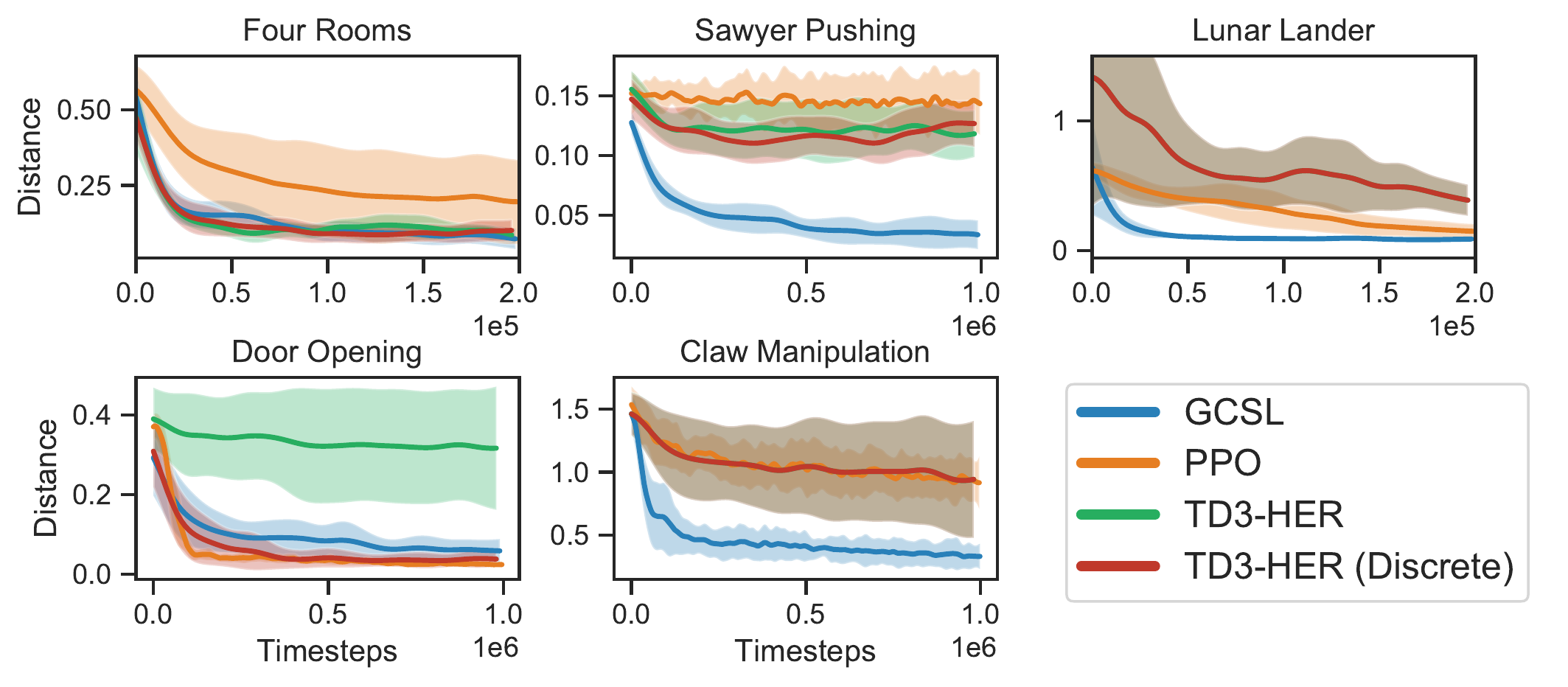}
    \includegraphics[width=0.8\linewidth]{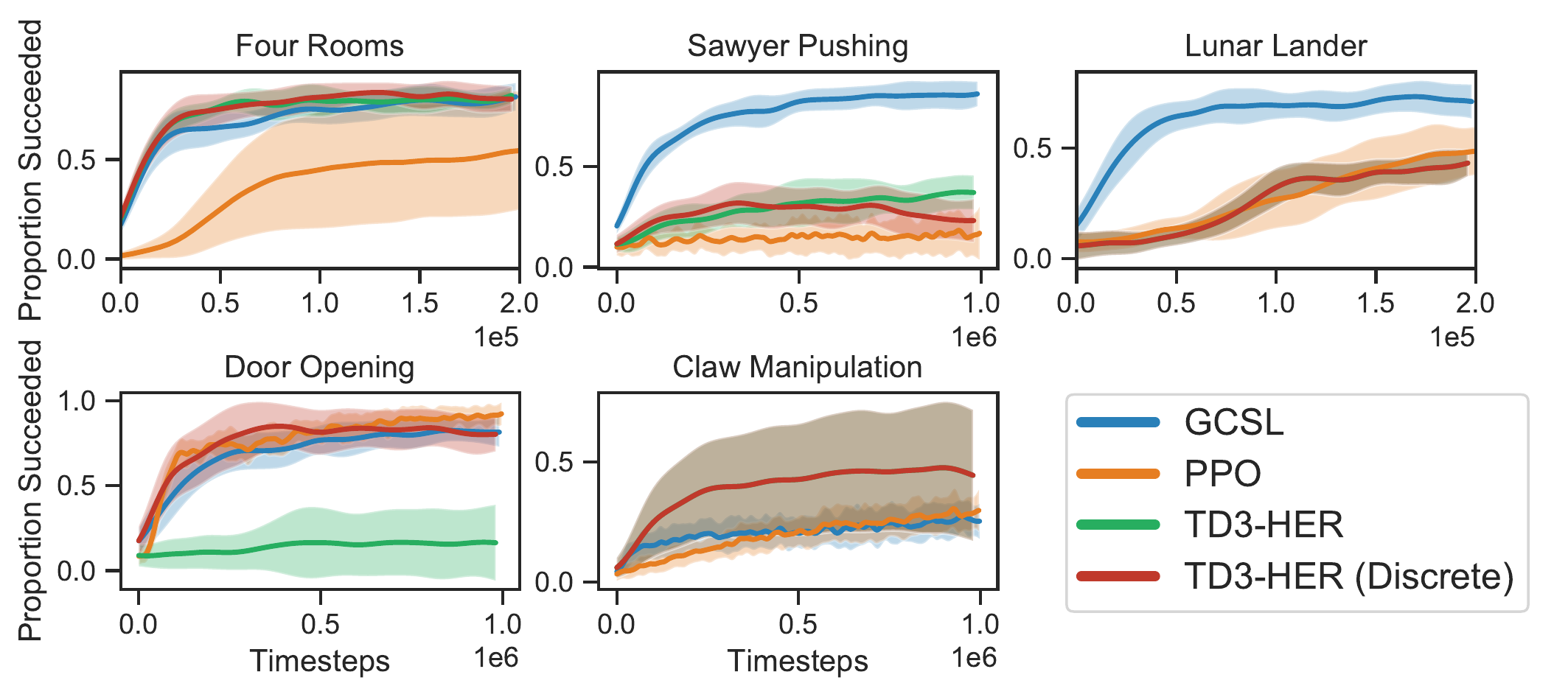}
    \caption{While we report results with median final distance in the main paper, here we present plots reporting both median final performance and proportion of trials that reached within $0.1m$ of the desired goal. This ``success ratio'' metric is well calibrated for all domains except for Claw manipulation, where final distance more accurately portrays the global performance of these algorithms. As mentioned in Appendix \ref{appendix:rl-details}, we additionally provide comparisons to a version of TD3-HER that uses discretized actions.}
    \label{fig:from-state-success}
\end{figure*}

We perform experimental comparisons with TD3-HER \citep{fujimoto2018off, andrychowicz2017her}. We relabel transitions as $((s,g), a, (s',g))$ gets relabeled to $((s,g'), a, (s',g'))$, where $g' = g$ with probability $0.1$, $g' = s'$ with probability $0.5$, and $g' = s_{t}$ for some future state in the trajectory $s_t$ with probability $0.4$. As described in Section $\ref{sec:background}$, the agent receives a reward of $1$ and the trajectory ends if the transition is relabeled to $g'=s'$, and $0$ otherwise. Under this formalism, the optimal $Q$-function, $Q^*(s,a,g) = \exp(-T(s,g))$, where $T(s,g)$ is the minimum expected time to go from $s$ to $g$. Both the Q-function and the actor for TD3 are parametrized as neural networks, with the same architecture (except final layers) for state-based domains as those for GCSL. We found the default values of learning rate, target update period, and number of critic updates to be the best amongst our hyperparameter search across the domains (single set of hyperparameters for all domains). Since GCSL uses discretized actions, we additionally compared to a version of TD3-HER that also uses discretized actions (results in Figure \ref{fig:from-state-success}). The performance of TD3-HER w/ discretized actions does not provide any increase on any environment except for the Door Opening task, where performance is competitive with GCSL.

We also compare to PPO \citep{schulman2017ppo}, an on-policy RL algorithm. Because PPO is on-policy, the algorithm cannot relabel goals; instead, we provide a surrogate $\epsilon$-ball indicator reward function: $r(s,g) = 1(d(s,g) < \epsilon)$, where $\epsilon$ is chosen appropriately for each environment. To maximize the data efficiency of PPO, we performed a coarse hyperparameter sweep over the batch size for the algorithm. Just as with TD3, we mimic the same neural network architecture for the parametrizations of the policies as GCSL. 

 \subsection{Task Descriptions}
\label{appendix:environment-details}

For each environment, the goal space is identical to the state space; each trajectory in the environment lasts for $50$ timesteps. 

\textbf{2D Room Navigation} \citep{ghosh2019learning}~ This environment requires an agent to navigate to points in an environment with four rooms that connect to adjacent rooms. The state space has two dimensions, consisting of the cartesian coordinates of the agent. The agent has acceleration control, and the action space has two dimensions. The distribution of goals $p(g)$ is uniform on the state space, and the agent starts in a fixed location in the bottom left room.

\textbf{Robotic Pushing} \citep{ghosh2019learning}~  This environment requires a Sawyer manipulator to move a freely moving block in an enclosed play area with dimensions $40$ cm $\times$ ~$20$ cm. The state space is $4$-dimensional, consisting of the Cartesian coordinates of the end-effector of the sawyer agent and the Cartesian coordinates of the block. The Sawyer is controlled via end-effector position control with a three-dimensional action space. The distribution of goals $p(g)$ is uniform on the state space (uniform block location and uniform end-effector location), and the agent starts with the block and end-effector both in the bottom-left corner of the play area.

\textbf{Lunar Lander} \citep{brockman2016gym}~ This environment requires a rocket to land in a specified region. The state space includes the normalized position of the rocket, the angle of the rocket, whether the legs of the rocket are touching the ground, and velocity information. Goals are sampled uniformly along the landing region, either touching the ground or hovering slightly above, with zero velocity.

\textbf{Door Opening:} \citep{nair2018rig} This environment requires a Sawyer manipulator to open a small cabinet door, initially shut closed, sitting on a table to a specified angle. The state space consists of the Cartesian coordinates of the Sawyer end-effector and the door's angle. As in the Robotic Pushing task, the three-dimensional action space controls the position of the end-effector. The distribution of goals $p(g)$ is uniform on door angles from 0 (completely closed) to 0.83 radians.

\textbf{Claw Manipulation:} \citep{ahn2019robel} A 9-DOF "claw"-like robot is required to turn a valve to various positions . The state space includes the positions of each joint of each claw (3 joints on 3 claws) and embeds the current angle of the valve in Cartesian coordinate ($\theta \mapsto (\sin \theta, \cos \theta)$). The robot is controlled via joint angle control. The goal space consists only of the claw angle, which is sampled uniformly from the unit circle.

\subsection{Ablations}
\label{appendix:ablation-details}

\begin{figure}[H]
    \centering
\includegraphics[width=0.19\linewidth]{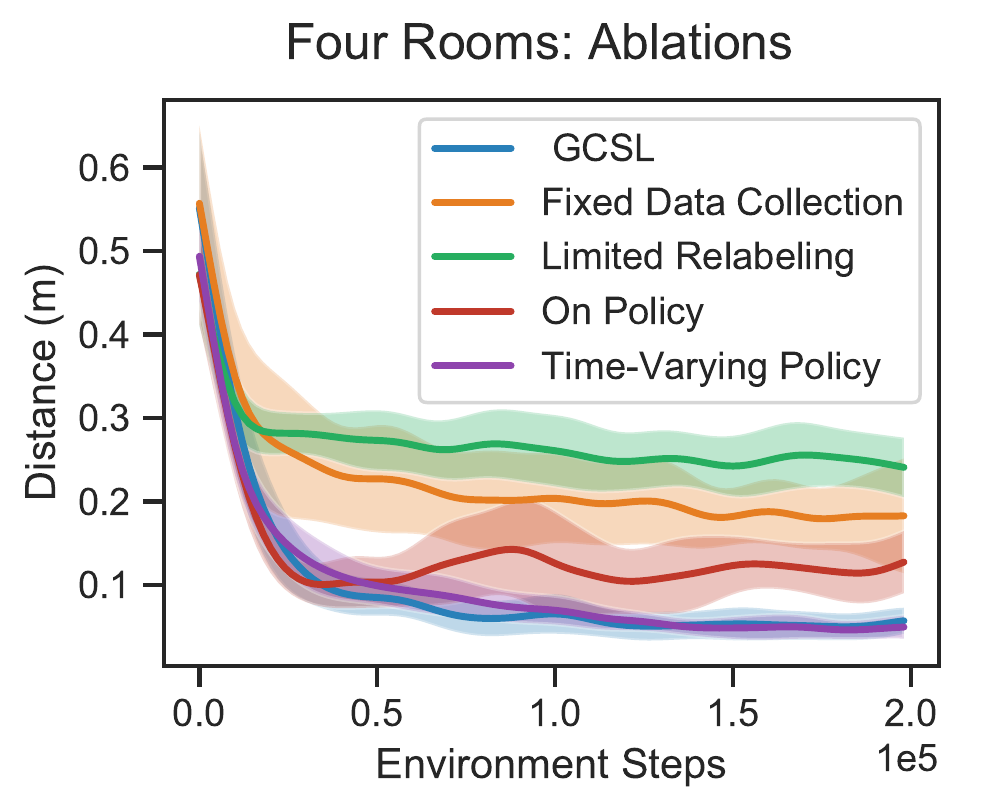}
    \includegraphics[width=0.19\linewidth]{plots/ablations/main/state_pusher_distance.pdf}
    \includegraphics[width=0.19\linewidth]{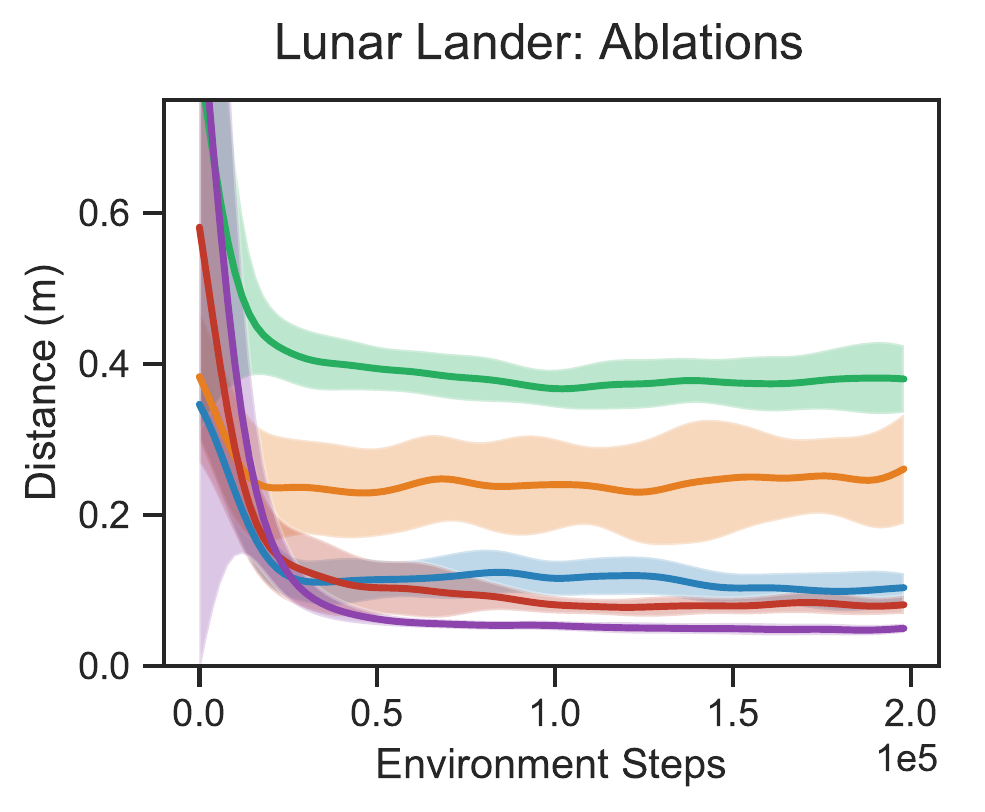}
    \includegraphics[width=0.19\linewidth]{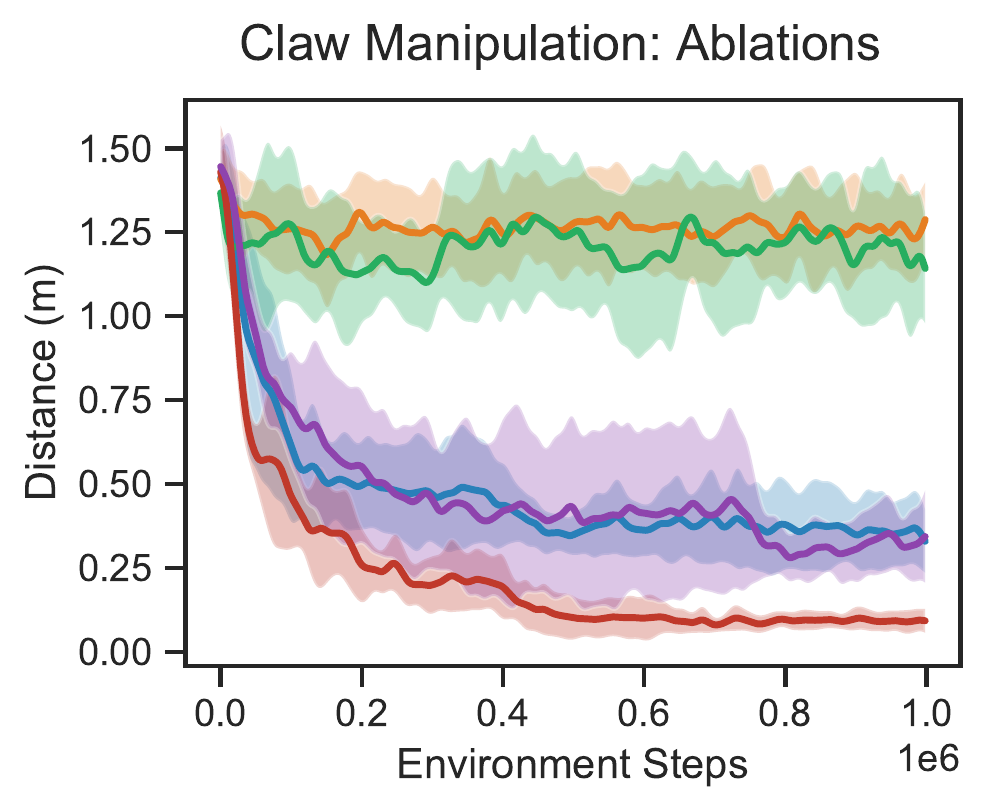}
    \includegraphics[width=0.19\linewidth]{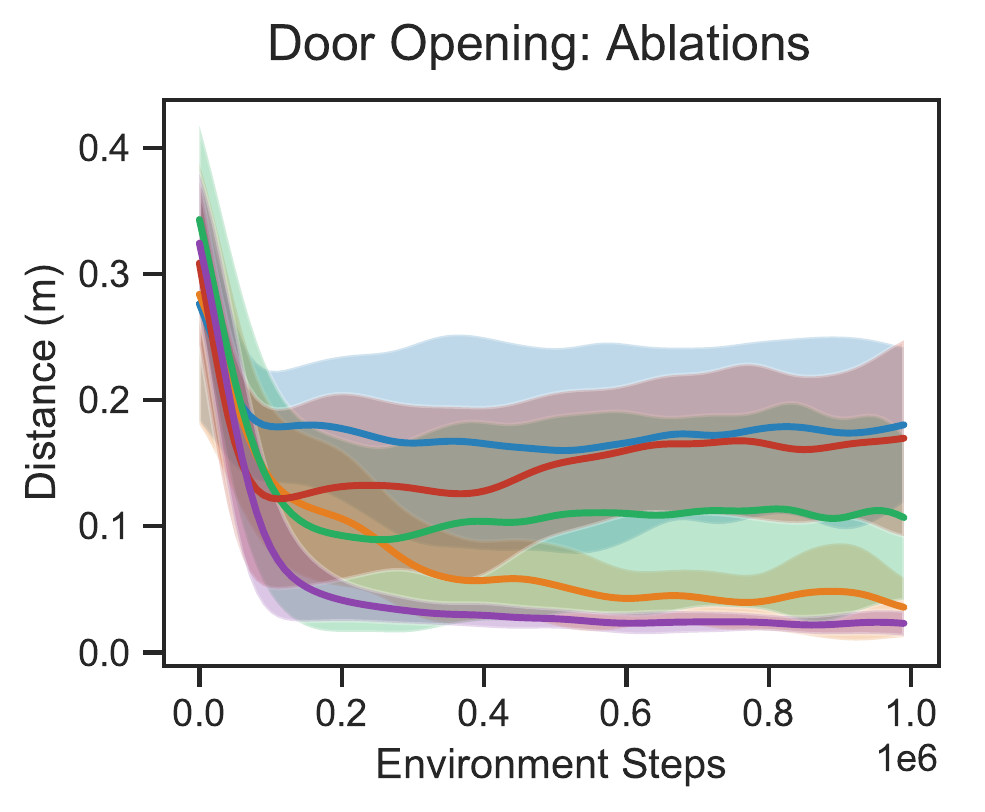}\\
    \includegraphics[width=0.19\linewidth]{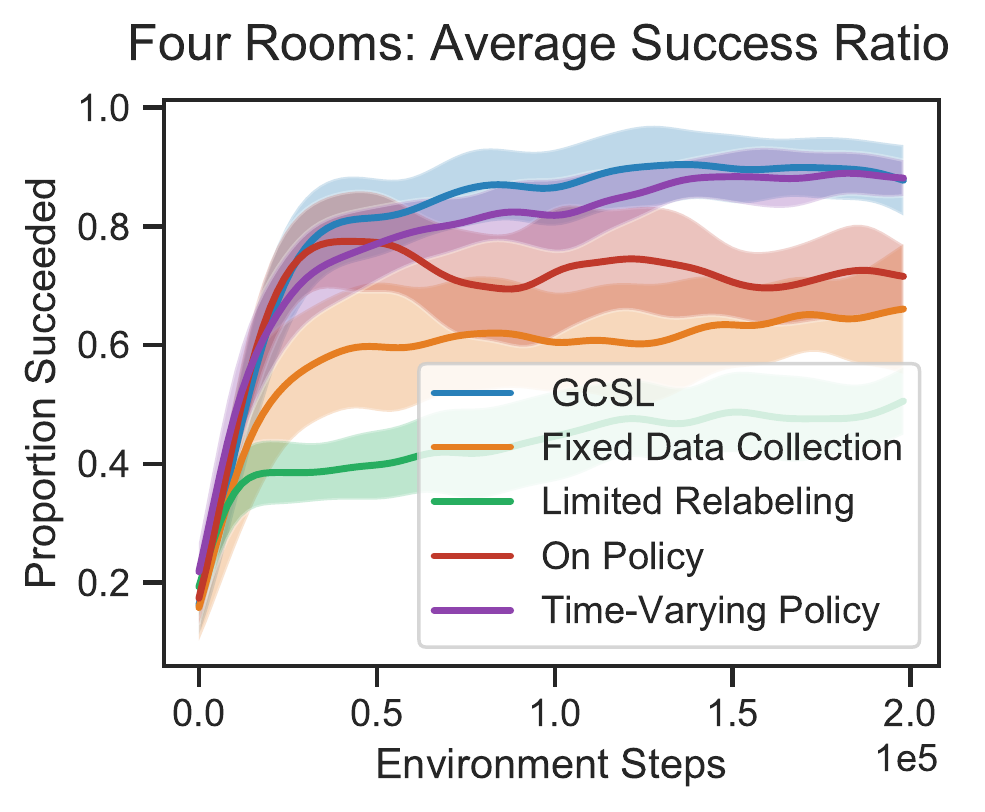}
    \includegraphics[width=0.19\linewidth]{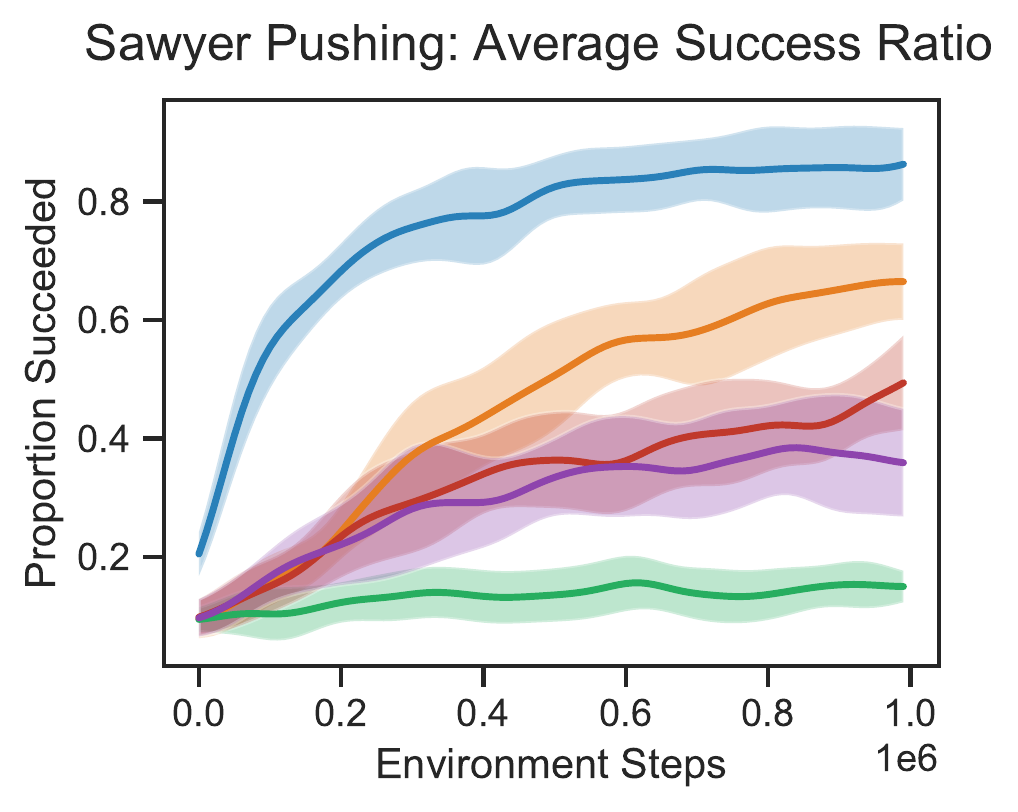}
    \includegraphics[width=0.19\linewidth]{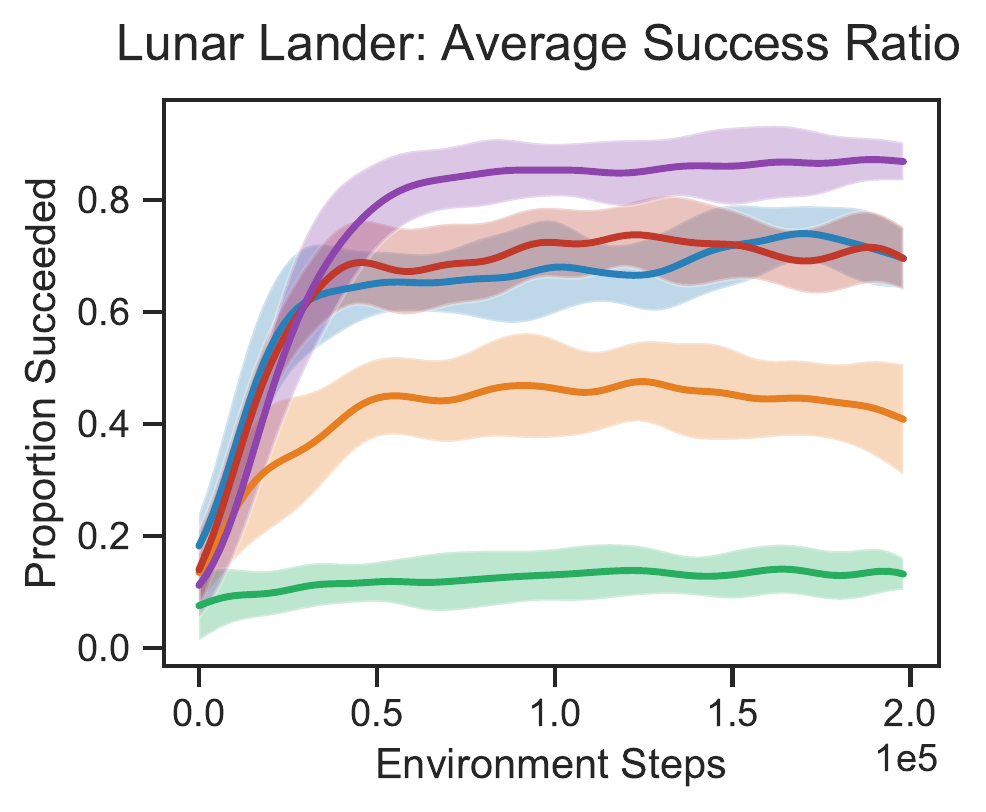}
    \includegraphics[width=0.19\linewidth]{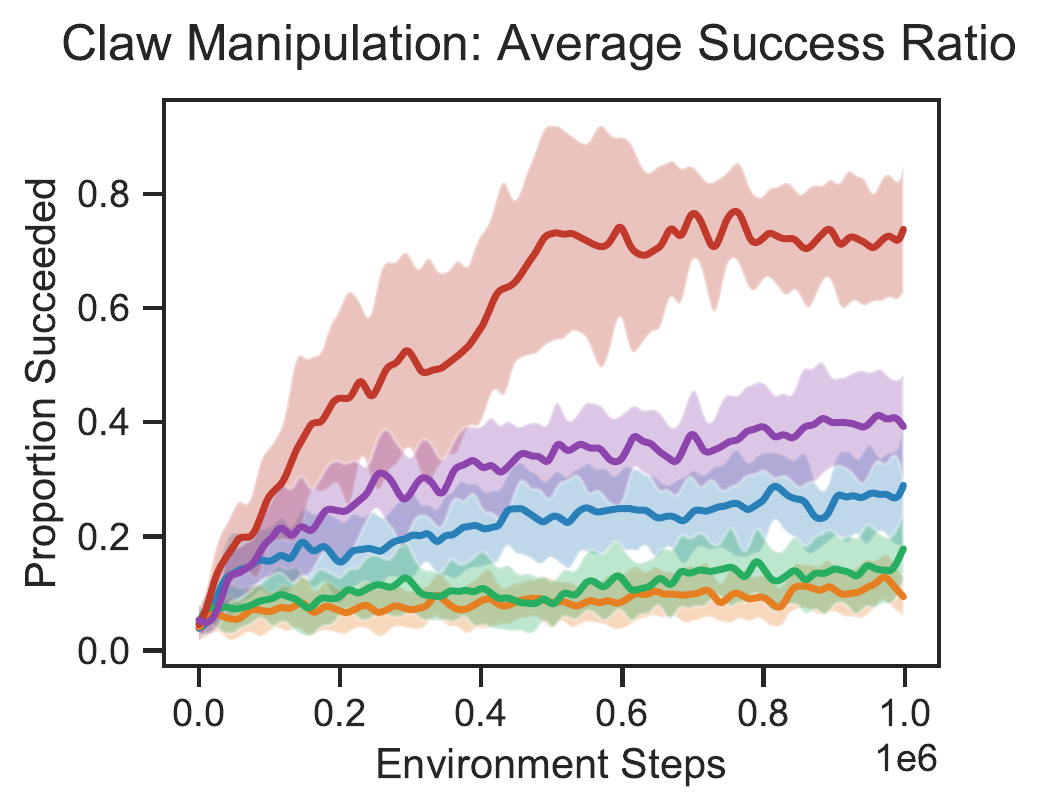}
    \includegraphics[width=0.19\linewidth]{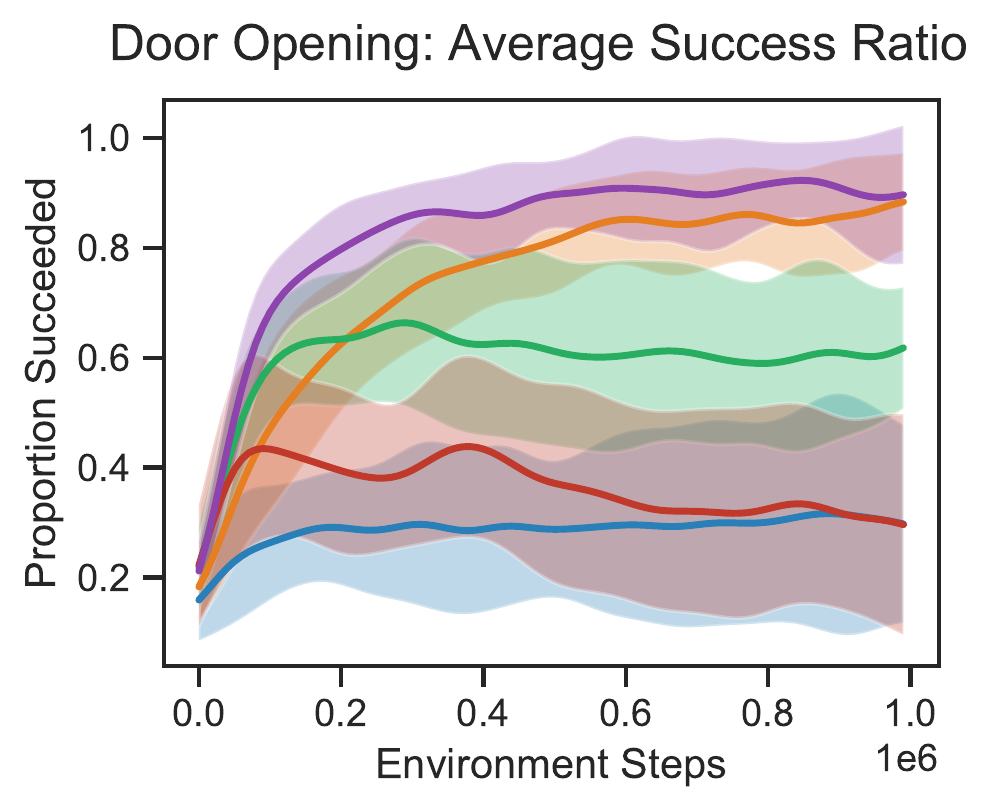}\\
    \caption{Performance across variations of GCSL (Section \ref{subsection:ablations}) for all experimental domains.} 
    
    \label{fig:appendix-ablations}
\end{figure}

In Section \ref{subsection:ablations}, we analyzed the performance of the following variants of GCSL (Figure \ref{fig:appendix-ablations}). 
\begin{enumerate}
    \item \textbf{Limited relabeling} - This model relabels only states and goals that are at most three steps apart: $\{(s_t, a_t, s_{t+h}, h) : t > 0, h \leq 3\}$
    \item \textbf{On-Policy} Only the most recent $10000$ transitions are stored and trained on.
    \item \textbf{Fixed Data Collection} Data is collected according to a uniform policy over actions.
    \item \textbf{Time-Varying Policy} Policies are are conditioned on the remaining horizon. Alongside the state and goal, the policy gets a reverse temperature encoding of the remaining horizon as input.
\end{enumerate}

\subsection{Hindsight Policy Gradients}
\label{appendix:hpg}
\begin{wrapfigure}{R}{0.4\textwidth}
\includegraphics[width=\linewidth]{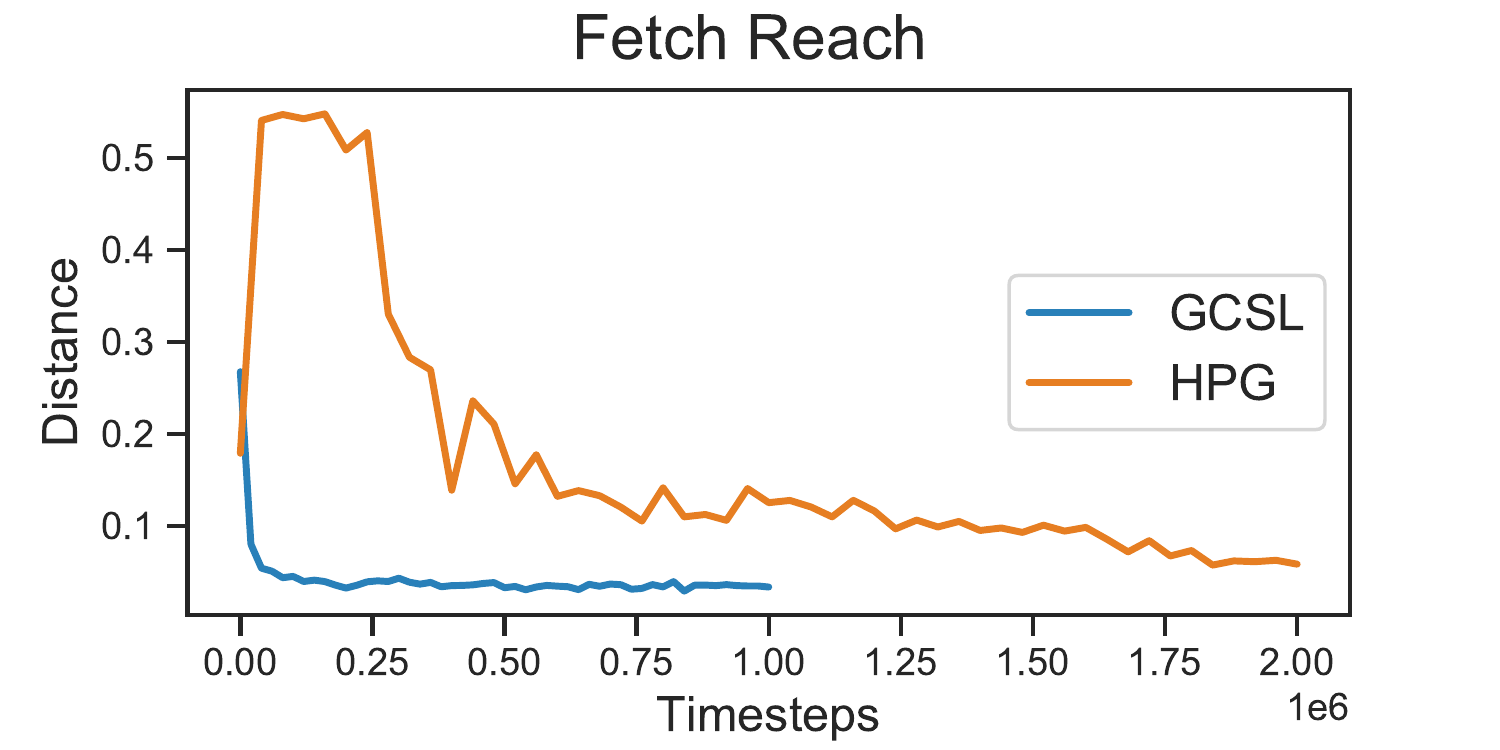}
\caption{{GCSL converges faster and learns a more accurate goal-reaching policy than HPG~\citep{rauber2017hindsight}\label{fig:hpg}}}
\end{wrapfigure}
We compared GCSL to hindsight policy gradients (HPG)~\citep{rauber2017hindsight} on the Fetch-Reach task from~\citet{andrychowicz2017her}. Whereas GCSL uses relabelling in conjunction with supervised imitation learning, HPG performs reward relabeling with a policy gradient algorithm, introducing importance weights to correct for the distribution shift.  As shown in Fig.~\ref{fig:hpg}, GCSL solves the task faster than HPG and converges to a policy that gets closer to the goal. HPG required more than 1 million samples to run reliably on Fetch-Reach, a task far easier to solve than any of our benchmark environments \citep{andrychowicz2017her}
\subsection{Initializing with Demonstrations}
\label{appendix:bootstrap-details}

We train an expert policy for robotic pushing using TRPO with a shaped dense reward function, and collect a dataset of 200 trajectories, each corresponding to a different goal. To train GCSL using these demonstrations, we simply populate the replay buffer with these trajectories at the beginning of training, and optimize the GCSL objective using these trajectories to warm-start the algorithm. Initializing a value function method using demonstrates requires significantly more attention: we perform the following procedure. First, we perform goal-conditioned behavior cloning to learn an initial policy $\pi_{BC}$. Next, we collect 200 new trajectories in the environment using a uniform data collection scheme. Using this dataset of $400$ trajectories, we perform policy evaluation on $\pi_{BC}$ to learn $Q^{\pi_{BC}}$ using policy evaluation via bootstrapping. Having trained such an estimate of the Q-function, we initialize the policy and Q-function to these estimates, and run the appropriate value function RL algorithm.
\section{Theoretical Analysis}

\subsection{Proof of Theorem 4.1}
\label{appendix:proof}

We will assume a discrete state space in this proof, and denote a trajectory as $\tau = \{s_0, a_0, \dots, s_T, a_T\}$.
Let the notation $\goalfn(\tau) = s_T$ denote the final state of a trajectory, which represents the goal that the trajectory reached. As there can be multiple paths to a goal, we let $\tau_g = \{\tau : \goalfn(\tau)=g\}$ denote the set of trajectories that reach a particular goal $g$. We abbreviate a policy's trajectory distribution as $\pi(\tau|g) = p(s_0)\prod_{t=0}^T \pi(a_t|s_t, g) \mathcal{T}(s_{t+1}|s_t,a_t)$. %
The target goal-reaching objective we wish to optimize is the probability of reaching a commanded goal, when goals are sampled from a pre-specified distribution $p(g)$.
\[
J(\pi) = \E_{g \sim p(g), \tau \sim \pi(\tau|g)}[\mathbbm{1}[\goalfn(\tau)=g]]
\]
GCSL optimizes the following objective, where the log-likelihood of the actions conditioned on the goals actually reached by the policy, $\goalfn(\tau)$. The distribution of trajectories used to optimize the objective is collected through a different policy, $\oldpi$. We write $\oldpi(\tau) = \E_{g \sim p(g)}[\oldpi(\tau|g)]$ to concisely represent the marginalized distribution of trajectories from $\oldpi$.

\begin{align*}
    J_{\text{GCSL}}(\pi) &= \E_{\tau \sim \oldpi(\tau)}\left[\sum_{t=0}^T \log \pi(a_t|s_t, \goalfn(\tau))\right]
\end{align*}
 To analyze how this objective relates to $J(\pi)$, we first analyze the relationship between $J(\pi)$ and a \emph{surrogate objective}, given by
\[
J_{\text{surr}}(\pi) = E_{g \sim p(g), \tau \sim \oldpi(\tau|g)}\left[\mathbbm{1}[\goalfn(\tau)=g] \log \pi (\tau|g)\right]
\]

Theorem 1 from \citet{schulman2015trpo} states that 
\[
J(\pi) \geq J_{\text{surr}}(\pi) - \frac{4\gamma\epsilon}{(1-\gamma)^2} \alpha^2,
\]
where $\gamma$ is a discount factor, $\epsilon$ is the maximum advantage over all states and actions, and $\alpha$ is the total variation distance between $\pi$ and $\oldpi$ . It is straightforward to show that the bound can be rewritten in the finite-horizon undiscounted case in terms of the horizon $T$, following \citet{Kakade:2002:AOA:645531.656005,dagger}, to obtain the bound
\[
J(\pi) \geq J_{\text{surr}}(\pi) - 4 T (T - 1) \epsilon \alpha^2,
\]
where $T$ is the horizon of the task. Since our reward function is $\mathbbm{1}[\goalfn(\tau)=g]$, the return for any trajectory is bounded between $0$ and $1$, allowing us to bound $\epsilon$ above by $1$. This leaves $\alpha$, which is the total variation divergence between $\pi$ and $\oldpi$. This divergence may be high if the data collection policy is very far from the current policy, but is low if the data was collected via a recent policy.

We can now lower-bound the surrogate objective with the GCSL objective via the following:
\begin{align*}
J_{\text{surr}}(\pi) &=  E_{g \sim p(g), \tau \sim \oldpi(\tau|g)}\left[\mathbbm{1}[\goalfn(\tau)=g] \log \pi (\tau|g)\right] \\
&= \sum_g p(g) \sum_\tau \oldpi(\tau|g) \log \pi(\tau| \goalfn(\tau))   \mathbbm{1}[\goalfn(\tau)=g]\\
&= \sum_\tau \log \pi(\tau| \goalfn(\tau)) \sum_g p(g) \oldpi(\tau|g) \mathbbm{1}[\goalfn(\tau)=g]\\
&= \sum_\tau \log \pi(\tau| \goalfn(\tau)) \sum_g p(g) \oldpi(\tau|g)
- \sum_\tau \log \pi(\tau| \goalfn(\tau)) \sum_g p(g) \oldpi(\tau|g) \mathbbm{1}[\goalfn(\tau) \neq g] \numberthis \label{analyze}\\
&\ge \sum_\tau \log \pi(\tau| \goalfn(\tau)) \sum_g p(g) \oldpi(\tau|g)\\
&= \E_{\tau \sim \E_g[\oldpi(\tau|g)]}[\log \pi(\tau|\goalfn(\tau))].\\
\end{align*}
The final line is our goal-relabeling objective: we train the policy to reach goals we reached. The inequality holds since $\log \pi(\tau)$ is always negative. The inequality is loose by a term related to the probability of not reaching the commanded goal, which we analyze in the section below.

Since the initial state and transition probabilities do not depend on the policy, we can simplify $\log \pi(\tau|\goalfn(\tau))$ as (by absorbing non $\pi$-dependent terms into $C_2$):
\begin{align*}
\E_{\tau \sim \oldpi(\tau)}[\log \pi(\tau|\goalfn(\tau))]
&= \E_{\tau \sim \oldpi(\tau)} \left [\log p(s_0)+\sum_{t=0}^T \log \pi(a_t|s_t,\goalfn(\tau))+\log \mathcal{T}(s_{t+1}|s_t,a_t) \right] \\
&= \E_{\tau \sim \oldpi(\tau)]} \left[\sum_{t=0}^T \log  \pi(a_t|s_t\goalfn(\tau)) \right]+C_2 \\
&= J_{\text{GCSL}}(\pi)+C_2.
\end{align*}
Combining this result with the bound on the expected return completes the proof:
\[J(\pi) \ge J_{GCSL}(\pi) + C_1+C_2 - 4T(T-1)\alpha^2\]

Note that in order for $J(\pi)$ and $J_{GCSL}(\pi)$ to be vacuously zero, the probability of reaching a goal under $\oldpi$ must be non-zero. This assumption is reasonable, and matches the assumptions on "exploratory data-collection" and full-support policies that are required by Q-learning and policy gradient convergence guarantees.

\subsection{Quantifying the Quality of the Approximation}
\label{appendix:proof-analysis}

The tightness of the bound presented above is controlled from two locations: the off-policyness of $\oldpi$ with respect to $\pi$ and the bound introduced by the lower bound in the theorem. The first is well-studied in policy gradient methods; in particular, when the data is on-policy, the gap between $J_{surr}(\pi)$ and $J(\pi)$ is known to be a policy-independent constant.  We seek to better understand the gap introduced by Equation \ref{analyze} in the analysis above. 

We define $P_{\oldpi}(\goalfn(\tau) \neq g)$ to be the probability of failure under $\oldpi$, and additionally define $p_{\text{wrong}}(\tau)$ and $p_{\text{right}}(\tau)$ to be the conditional distribution of trajectories under $\oldpi$ given that it did not reach and did the commanded goal respectively. 

In the following section, we show that the gap introduced by Equation \ref{analyze} can be controlled by the probability of making a mistake, $P_{\oldpi}(\goalfn(\tau) \neq g)$, and $D_{TV}(p_{\text{wrong}}(\tau), p_{\text{right}}(\tau))$, a measure of the difference between the distribution of trajectories that must be relabeled and those not.

We rewrite Equation \ref{analyze} as follows: 
\begin{align*}
J_{surr}(\pi) &= \sum_\tau \log \pi(\tau| \goalfn(\tau)) \sum_g p(g) \oldpi(\tau|g)
- \sum_\tau \log \pi(\tau| \goalfn(\tau)) \sum_g p(g) \oldpi(\tau|g) \mathbbm{1}[\goalfn(\tau) \neq g] \\ 
&= \E_{\tau \sim \oldpi}[\log \pi(\tau|\goalfn(\tau))]
 - P_{\oldpi}\left(\goalfn(\tau) \neq g) \right)\mathbb{E}_{\tau \sim p_{\text{wrong}}(\tau)}\left[\log \pi(\tau | \goalfn(\tau))\right]  \\
 \intertext{Define $D$ to be the Radon-Nikodym derivative of $p_{\text{wrong}}(\tau)$ wrt $\oldpi(\tau)$ }
 &= \E_{\tau \sim \oldpi(\tau)}[\log \pi(\tau|\goalfn(\tau))]
 - P_{\oldpi}\left(\goalfn(\tau) \neq g) \right)\mathbb{E}_{\tau \sim \oldpi(\tau)}\left[D \log \pi(\tau | \goalfn(\tau))\right]  \\
  &= (1 - P_{\oldpi}\left(\goalfn(\tau) \neq g\right)) \E_{\tau \sim \oldpi(\tau)}[\log \pi(\tau|\goalfn(\tau))]\\
 &~~~~~+ \underbrace{P_{\oldpi}\left(\goalfn(\tau) \neq g) \right)\mathbb{E}_{\tau \sim \oldpi(\tau)}\left[(1 - D) \log \pi(\tau | \goalfn(\tau))\right]}_{\text{Relevant Gap}}  \\
\end{align*}
The first term is affine with respect to the GCSL loss, so the second term is the error we seek to understand. 
\begin{align*}
|\text{Relevant Gap}| &= P_{\oldpi}\left(\goalfn(\tau) \neq g\right)\left|\mathbb{E}_{\tau \sim \oldpi(\tau)}\left[(1 - D) \log \pi(\tau | \goalfn(\tau))\right]\right|\\
&\leq P_{\oldpi}(\goalfn(\tau) \neq g) \mathbb{E}_{\tau \sim \oldpi}[|1 - D|] \mathbb{E}_{\tau \sim \oldpi(\tau)}[\log \pi(\tau | \goalfn(\tau))]\\
&= 2 P_{\oldpi}(\goalfn(\tau) \neq g) D_{TV}\left(\E_g[\oldpi(\tau|g)], p_{\text{wrong}}(\tau)\right) \mathbb{E}_{\tau \sim \oldpi(\tau)}[\log \pi(\tau | \goalfn(\tau))]\\
&= 2 P_{\oldpi}(\goalfn(\tau) \neq g) (1 - P_{\oldpi}(\goalfn(\tau) \neq g)) D_{TV}\left(p_{\text{right}}(\tau), p_{\text{wrong}}(\tau)\right) \mathbb{E}_{\tau \sim \oldpi(\tau)}[\log \pi(\tau | \goalfn(\tau))]\\
\end{align*}

The inequality is maintained because of the nonpositivity of $\log \pi(\tau)$, and the final step holds because $\oldpi(\tau)$ is a mixture of $p_{\text{wrong}}(\tau)$ and $p_{\text{right}}(\tau)$. This derivation shows that the gap between $J_{surr}$ and $J_{GCSL}$ (up to affine consideration) can be controlled by (1) the probability of reaching the wrong goal and (2) the divergence between the conditional distribution of trajectories which did reach the commanded goal (do not need to be relabeled) and those which did not reach the commanded goal (must be relabeled). As either term goes to $0$, this bound becomes tight.
\subsection{Proof of Theorem \ref{thm:performance_main}}
\label{appendix:proof-performance}
In this section, we now prove that sufficiently optimizing the GCSL objective over the full state space causes the probability of reaching the wrong goal to be bounded close to $0$, and thus bounds the gap close to $0$.

Suppose we collect trajectories from a policy $\oldpi$. Following the notation from before, we define $\oldpi(\tau) = \mathbb{E}_{g \sim p(g)}[\pi_{data}(\tau | g)]$. For convenience, we define $\pi^*(a_t | s_t, g) \propto \int_{\tau \setminus a_t} \pi_{data}(\tau)1(\goalfn(\tau) = g)1(s_t(\tau) = s_t)$ to be the conditional distribution of actions for a given state given that the goal $g$ is reached at the end of the trajectory. If this conditional distribution is not defined, we let $\pi^*(a_t | s_t, g)$ be uniform, so that $\pi^*(a_t | s_t, g)$ is well-defined for all states, goals, and timesteps. The notation for $\pi^*$ is suggestive: in fact, it can be easily shown that under the assumptions of the theorem, full data coverage and deterministic dynamics, the induced policy $\pi^*$ is in fact the optimal policy for maximizing the probability of reaching the goal.

To show that the GCSL policy also incurs low error, we provide a coupling argument, similar to \cite{schulman2015trpo, Kakade:2002:AOA:645531.656005, dagger}. Because $D_{TV}(\pi(a_t | s_t, g), \pi^*(a_t | s_t, g)) \leq \epsilon$, we can define a $(1-\epsilon)$-coupled policy pair $(\overline{\pi}, \overline{\pi^*})$, which take differing actions with probability $\epsilon$. By a union bound over all timesteps, the probability that $\overline{\pi}$ and $\overline{\pi^*}$ take any different actions throughout the trajectory is bounded by $\epsilon T$, and because of the assumptions of deterministic dynamics, take the same trajectory with probability $1-\epsilon T$. 
Now, since the two policies take different trajectories with probability \textit{at most }$\epsilon T$, a simple bound shows that the probability that $\pi_{GCSL}$ reaches the goal is at most $\epsilon T$ less than $\pi*$, leading to our result that the performance gap $J(\pi^*) - J(\pi) < \epsilon T$. In environments in which every state is reachable from every other state in the desired horizon, this provides a global performance bound indicating that the optimal GCSL policy will reach the goal with probability at least $1-\epsilon T$.

\section{Example Trajectories}
\label{appendix:example_trajectories}
Figure \ref{fig:example-trajectories} below shows parts of the state along trajectories produced by GCSL. In Lunar Lander, this state is captured by the rocket's position, and in 2D Room Navigation it is the agent's position. While these trajectories do not always take the shortest path to the goal, they do often take fairly direct paths to the goal from the initial position avoiding very roundabout trajectories. 

\begin{figure}[H]
    \centering
\includegraphics[width=0.4\linewidth]{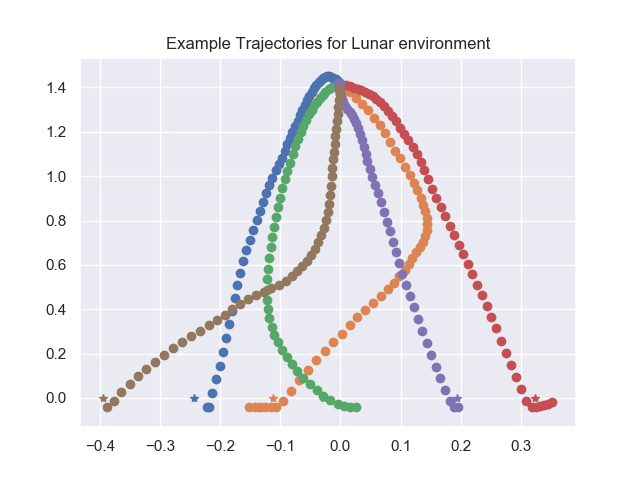}
    \includegraphics[width=0.4\linewidth]{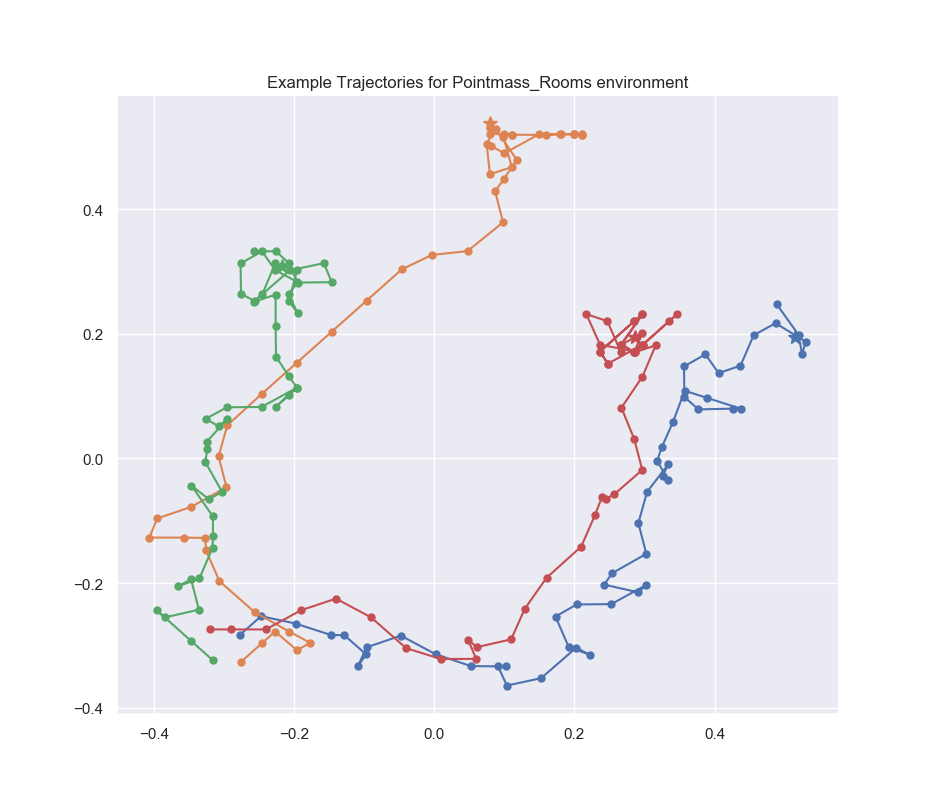}
    \caption{Examples of trajectories generated by GCSL for the Lunar Lander and 2D Room environments. Stars indicate the goal state.} 
    \label{fig:example-trajectories}
\end{figure}

\end{document}